\pdfoutput=1
\def\year{2021}\relax
\documentclass[letterpaper]{article} 
\usepackage{aaai21}  
\usepackage{times}  
\usepackage{helvet} 
\usepackage{courier}  
\usepackage[hyphens]{url}  
\usepackage{graphicx} 
\usepackage{natbib}  
\usepackage{caption} 
\usepackage{bm}
\usepackage{amsthm}
\usepackage{amsmath, amssymb, amsfonts}
\usepackage{algorithm}
\usepackage{algorithmic}
\usepackage{makecell}
\usepackage{paralist}
\usepackage[american]{babel}
\usepackage{subfigure}

\newcommand{\x}{\textbf{x}}

\newtheorem{theorem}{Theorem}

\DeclareMathOperator*{\argmin}{argmin}
\theoremstyle{definition}
\newtheorem{remark}{Remark}

\title{Storage Fit Learning with Feature Evolvable Streams\thanks{This research was supported by NSFC (61921006).}}

\author{
Bo-Jian Hou, Yu-Hu Yan, Peng Zhao, Zhi-Hua Zhou\\
}
\affiliations{
National Key Laboratory for Novel Software Technology, \\
Nanjing University, Nanjing 210023, China\\
\{houbj, yanyh, zhaop, zhouzh\}@lamda.nju.edu.cn
}

\begin{document}

\maketitle

\begin{abstract}
    Feature evolvable learning has been widely studied in recent years where old features will vanish and new features will emerge when learning with streams. Conventional methods usually assume that a label will be revealed after prediction at each time step. However, in practice, this assumption may not hold whereas no label will be given at most time steps. A good solution is to leverage the technique of manifold regularization to utilize the previous similar data to assist the refinement of the online model. Nevertheless, this approach needs to store all previous data which is impossible in learning with streams that arrive sequentially in large volume. Thus we need a buffer to store part of them. Considering that different devices may have different storage budgets, the learning approaches should be flexible subject to the storage budget limit. In this paper, we propose a new setting: \emph{Storage-Fit Feature-Evolvable streaming Learning} (SF$^2$EL) which incorporates the issue of rarely-provided labels into feature evolution. Our framework is able to fit its behavior for different storage budgets when learning with feature evolvable streams with unlabeled data. Besides, both theoretical and empirical results validate that our approach can preserve the merit of the original feature evolvable learning i.e., can always track the best baseline and thus perform well at any time step. 
    
\end{abstract}

\section{Introduction}
\label{section:introduction}
Over the last several years, \emph{feature evolvable learning} has drawn extensive attentions~\cite{DBLP:journals/tkde/ZhangZL0ZW16,DBLP:conf/nips/Hou0Z17,DBLP:journals/pami/HouZ18,DBLP:conf/icml/YeZ0Z18,DBLP:conf/icml/Zhang0JZ20}, where old features will vanish and new features will emerge when data streams come continuously or in an online manner. There are various problem settings proposed in previous studies. For instance, in FESL~\cite{DBLP:conf/nips/Hou0Z17}, there is an \emph{overlapping period} where old and new features exist simultaneously when feature space switches. \citet{DBLP:journals/pami/HouZ18} investigate the scenario when old features disappear, part of them will survive and continue to exist with new arriving features. \citet{DBLP:journals/tkde/ZhangZL0ZW16} study that features of new samples are always equal to or larger than the old samples so as to render \emph{trapezoidal data streams}. Subsequent works consider the situation that features could vary arbitrarily at different time steps under certain assumptions~\cite{DBLP:conf/aaai/BeyazitA019,DBLP:conf/ijcai/HeWWBC019}. 

Note that the setting of feature evolvable learning is different from transfer learning~\cite{DBLP:journals/tkde/PanY10} or domain adaptation~\cite{jiang2008literature,DBLP:journals/inffus/SunSW15}. Transfer learning usually assumes that data come in batches instead of the streaming form. One exception is online transfer learning~\cite{DBLP:journals/ai/ZhaoHWL14} in which data from both sets of features arrive sequentially. However, they assume that all the feature spaces must appear simultaneously during the whole learning process while such an assumption does not hold in feature evolvable learning. Domain adaptation usually assumes the data distribution changes across the domains, yet the feature spaces are the same, which is evidently different from the setting of feature evolvable learning.

These conventional feature evolvable learning methods all assume that a label can be revealed in each round. However, in real applications, labels may be rarely given during the whole learning process. For example, in an object detecting system, a robot takes high-frame rate pictures to learn the name of different objects. Like a child learning in real world, the robot receives names rarely from human. Thus we will face the \emph{online semi-supervised learning} problem. We focus on manifold regularization which assumes that similar samples should have the same label and has been successfully applied in many practical tasks~\cite{zhu2005semi}. However, this method needs to store previous samples and render a challenge on storage. Besides, different devices have different storage budget limitations, or even the available storage in the same device could be different at different times. Thus it is important to make our method adjust its behavior to fit for different storage budgets (known as {\it storage-fit issues}) ~\cite{DBLP:conf/pakdd/ZhouNSJ09,DBLP:conf/ijcai/Hou0Z17} which means the method should fully exploit the storage budget to optimize its performance.

In this paper, we propose a new setting: \emph{Storage-Fit Feature-Evolvable streaming Learning} (SF$^2$EL) which concerns both the lack of labels and the storage-fit issue in the feature evolvable learning scenario. We focus on FESL~\cite{DBLP:conf/nips/Hou0Z17}, and other feature evolvable learning methods based on online learning technique can also adapt to our framework since our framework is not affected by specific forms of feature evolution. Due to the lack of labels, the loss function of FESL cannot be used anymore. We leverage manifold regularization technique to convert its loss function to a ``risk function" without the need to consider if there exists a label or not. We also make use of a buffer strategy named {\it reservoir sampling}~\cite{DBLP:journals/toms/Vitter85} when encountering the storage-fit problem. Our contributions are threefold as follows.
\begin{itemize}
\item Both theoretical and experimental results demonstrate that our method is able to always follow the best baseline at any time step. Thus our model can always perform well during the whole learning process in new feature space regardless of the limitation that only few data emerge in the beginning. This is a very fundamental requirement in feature evolvable learning scenario and FESL as well as other feature evolvable learning methods cannot achieve this goal when labels are barely given.
\item In addition, the experimental results indicate that manifold regularization plays an important role when there are only few labels. 
\item Finally, we theoretically and experimentally validate that larger buffer brings better performance. Therefore, our method can fit for different storages by taking full advantage of the budget.
\end{itemize}

The rest of this paper is organized as follows. Section~2 introduces the preliminary about the basic scenario on feature evolution and the framework of our approach. Our proposed approach with two corresponding analyses is presented in Section~3. Section~4 reports experimental results. Finally, Section~5 concludes our paper.

\section{Preliminary and Framework}
\label{section:preliminary}

We focus on binary classification task. On each round of the learning process, the learner observes an instance and gives its prediction. After the prediction has been made, with a small probability $p_l$, the true label is revealed. Otherwise, the instance remains unlabeled. The learner updates its predictor based on the observed instance and the label, if any. In the following, we introduce FESL that we are interested in. 

\begin{figure}[!t]
    \centering
    \includegraphics[width=1\linewidth]{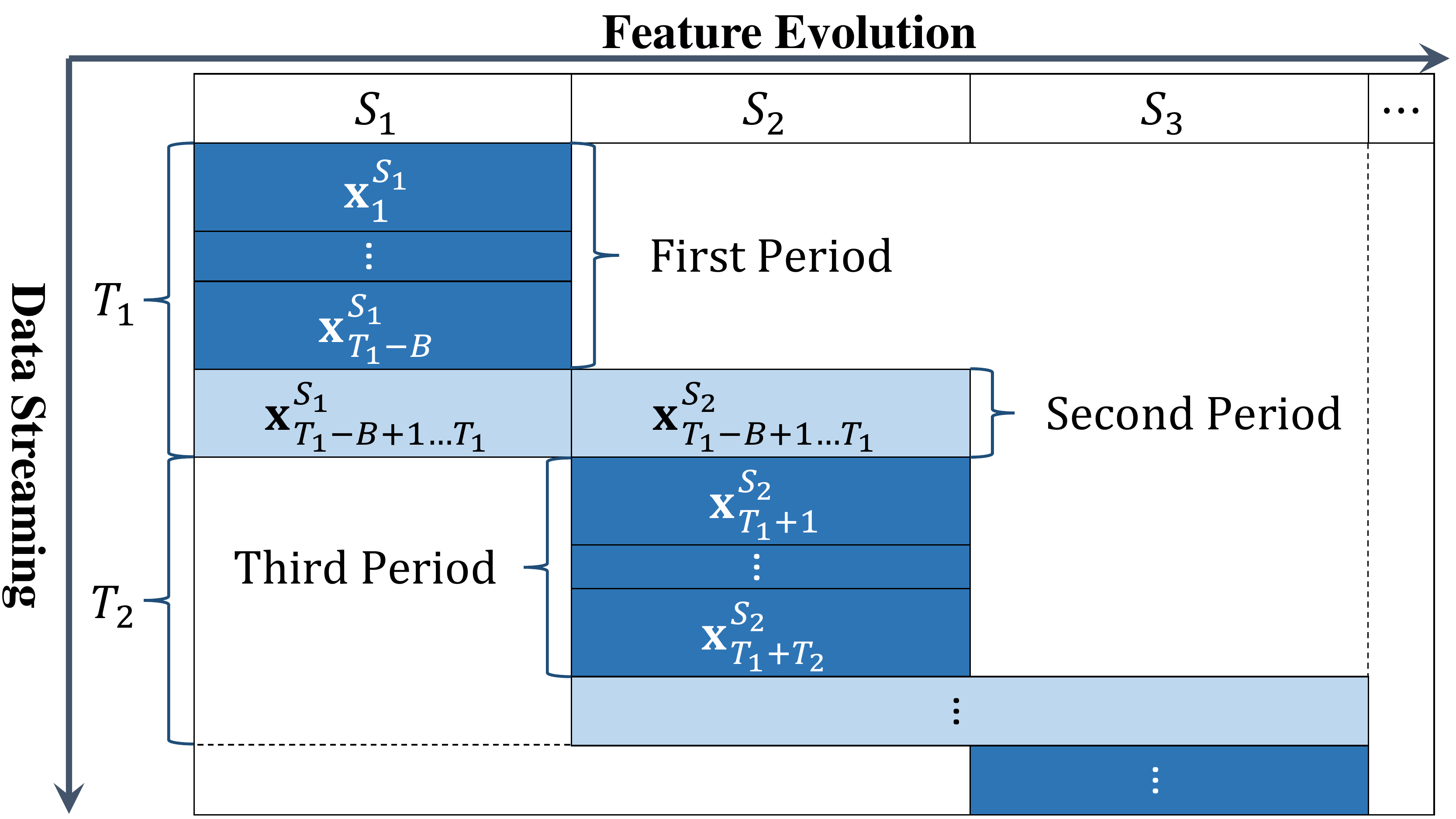}
    \caption{Illustration of how data stream comes.}
    \label{illustration-general}
    \vspace{-0.2cm}
\end{figure}

FESL defines ``feature space" by a set of features. And ``the feature space changes" means both the underlying distribution of the feature set and the number of features change. Figure~\ref{illustration-general} illustrates how data stream comes in FESL.  There are three repeating periods: in the first period a large amount of data streams come from the old feature space $S_1$; then in the second period named ``overlapping period", few of data come from both the old and the new feature space, i.e., $S_1$ and $S_2$; soon afterwards in the third period, data streams only come from the new feature space $S_2$. These three periods will continue again and again and form cycles. Each cycle merely includes two feature spaces and thus, we only need to focus on one cycle and it is easy to extend to the case with multiple cycles. Besides, FESL assumes that the old features in one cycle will vanish simultaneously according to the example of ecosystem protection where all the sensors share the same expected lifespan and thus they will wear out at the same time. The case where old features vanish asynchronously has been studied in PUFE~\cite{DBLP:journals/corr/abs-1904-12171}, which can adapt to our framework as well.

Based on the above discussion, we only consider two feature spaces denoted by $S_1$ and $S_2$, respectively. Suppose that in the overlapping period, there are $B$ rounds of instances both from $S_1$ and $S_2$. As can be seen from Figure~\ref{illustration-general}, the process can be summarized as follows.

\begin{itemize}
\item For $t=1,\ldots,T_1-B$, in each round, the learner observes a vector $\x_t^{S_1}\in\mathbb{R}^{d_1}$ sampled from $S_1$ where $d_1$ is the number of features of $S_1$, $T_1$ is the number of total rounds in $S_1$.
\item For $t=T_1-B+1,\ldots,T_1$, in each round, the learner observes two vectors $\x_t^{S_1}\in\mathbb{R}^{d_1}$ and $\x_t^{S_2}\in\mathbb{R}^{d_2}$ from $S_1$ and $S_2$ where $d_2$ is the number of features of $S_2$.
\item For $t=T_1+1,\ldots,T_1+T_2$, in each round, the learner observes a vector $\x_t^{S_2}\in\mathbb{R}^{d_2}$ sampled from $S_2$ where $T_2$ is the number of rounds in $S_2$. Note that $B$ is small, so we can omit the streaming data from $S_2$ on rounds $T_1-B+1,\ldots,T_1$ since they have minor effect on training the model in $S_2$.
\end{itemize}


FESL adopts linear predictor, whereas to be general, nonlinear predictor is chosen in our paper. Let $K_i$ denote a kernel over $\x^{S_i}$ and $\mathcal{H}_{K_i}$ the corresponding Reproducing Kernel Hilbert Space (RKHS)~\cite{DBLP:books/lib/ScholkopfS02} where $i=1,2$ that indexes the feature space. We define the projection as $\Pi_{\mathcal{H}_{K}}(b)=\argmin_{\textbf{a}\in\mathcal{H}_{K}}\|a-b\|_{K}$. The predictor learned from the sequence is denoted as $f\in\mathcal{H}_K$. Denote by $f_{i,t},\,i=1,2$ the predictor learned from the $i$th feature space in the $t$th round. The loss function $\ell(f(\x),y)$ is convex in its first argument such as {\it logistic loss}
$
\ell(f(\x),y)=\ln(1+\exp(-yf(\x))),
$
{\it hinge loss}
$
\ell(f(\x),y)=\max(0,1-yf(\x)),
$
etc., for classification tasks.

If the label is fully provided, the risk suffered by the predictor in each round can be merely the prediction loss mentioned above. Then the most straightforward or baseline algorithm is to apply online gradient descent~\cite{DBLP:conf/icml/Zinkevich03} on rounds $1,\ldots,T_1$ with streaming data $\x_t^{S_1}$, and invoke it again on rounds $T_1+1,\ldots,T_1+T_2$ with streaming data $\x_t^{S_2}$. The models are updated according to: 
\begin{equation}
\label{equation:update model}
f_{i,t+1}=\Pi_{\mathcal{H}_{K_i}}\left(f_{i,t}-\tau_t\nabla\ell(f_{i,t}(\x_t^{S_i}),y_t)\right),\,i=1,2,
\end{equation}
where $\nabla\ell(f_{i,t}(\x_t^{S_i}),y_t)$ is the gradient of the loss function on $f_{i,t}$ and $\tau_t$ is a time-varying step size, e.g., $\tau_t=1/\sqrt{t}$. 

Nevertheless, the fundamental goal of FESL and other feature evolvable learning methods is that the model can always keep the performance at a good level no matter in the beginning of each feature space or at any other time. This baseline method cannot achieve this goal since there are only few data in the beginning of $S_2$ and it is difficult to obtain good performance with only training on these few data. A fundamental idea of FESL and other feature evolvable learning methods is to establish a relationship between the old feature space and the new one. In this way, the learning on the new feature space can be assisted by the old well-learned model and the goal can be achieved. 

Specifically, FESL learns a mapping $\psi:\mathbb{R}^{d_2}\rightarrow\mathbb{R}^{d_1}$ between $S_1$ and $S_2$ by least squares during the overlapping period. Then when $S_1$ disappears, we can leverage this mapping to map the new data from $S_2$ into $S_1$ to recover the data from $S_1$, i.e., $\psi(\x_t^{S_2})$. At this rate, the well-learned model $f_{1,T_1}$ from $S_1$ can make good prediction on the recovered data $\psi(\x_t^{S_2})$ and update itself with them. Concurrently, a new model is learned in $S_2$ and another prediction on $\x_t^{S_2}$ is also made. At the beginning, the $S_1$'s prediction $f_{1,t}(\psi(\x_t^{S_2}))$ is good with the good predictor $f_{1,t}$ and $S_2$'s prediction $f_{2,t}(\x_t^{S_2})$ is bad due to limited data. But after some time, $f_{1,t}(\psi(\x_t^{S_2}))$ may become worse because of the cumulated error brought by the inaccurate mapping and $f_{2,t}(\x_t^{S_2})$ will be better with more and more accurate data. FESL dynamically combines these two changing predictions with weights by calculating the loss of each base model. With this strategy, it achieves the \emph{fundamental goal} in feature evolvable learning, i.e., can always follow the best base model at any time step and thus always perform well during the whole learning process in the new feature space.

Unfortunately, however, we cannot always obtain a label in each round. Thus (\ref{equation:update model}) cannot be calculated so that FESL and other feature evolvable learning methods cannot achieve the goal. We leverage manifold regularization technique to mitigate this problem such that we can continue to calculate our risk function even when no labels are provided. But this operation requires the calculation of similarity between each observed sample and the current sample. This brings huge burdens on the storage and computation, which is not allowed in streaming learning or online learning scenario. Therefore, we incorporate the buffering strategy which only uses a small buffer to store representative samples. Considering that different devices provide different storage budgets, and even the same device will provide different available storages, we need to fit our method for different storages to maximize the performance (known as storage-fit issue), which can be accomplished by our buffering strategy. 

So far, our framework is clear, that is: 
\begin{itemize}
\item We first exploit manifold regularization to mitigate the problem where labels are rarely given, and then the online gradient descent can be calculated again;
\item Based on the modification in the first step, we can derive our learning procedure from FESL naturally, yet with a potential problem of storage and computation;
\item Finally, we use a buffering strategy to solve the storage and computation problem and subsequently solve the storage-fit issue based on this strategy.
\end{itemize}


\section{Our Approach}
\label{section:our approach}
Based on the framework described in the end of the last section, in this section, we introduce our approach along the way of considering ``manifold regularization'', ``combining base learners'' and ``buffering''. In the end of this section, we also provide two analyses with respect to the fundamental goal and the storage-fit issue respectively.

\subsection{Manifold Regularization}
With limited labels, we will face an \emph{online semi-supervised learning} problem. There are several convex semi-supervised learning methods, e.g., manifold regularization and multi-view learning. Their batch risk is the sum of \emph{convex function} in $f$. For these convex semi-supervised learning methods, one can derive a corresponding online semi-supervised learning algorithm using online convex programming~\cite{DBLP:conf/pkdd/GoldbergLZ08}. We focus on manifold regularization while the online versions of multi-view learning and other convex semi-supervised learning methods can be derived similarly.

In online learning, the learner only has access to the input sequence up to the current time. We thus define the \emph{instantaneous regularized risk} $J_{i,t}(f_{i,t})$ at time $t$ to be 
\begin{equation}
    \setlength{\abovedisplayskip}{1pt}
    \setlength{\belowdisplayskip}{1pt}
    \label{equation:instan risk}
    \begin{split}
    J_{i,t}(f_{i,t})=\frac{T}{l}\delta(y_t)\ell(f_{i,t}(\x_t^{S_i}),y_t)+\frac{\lambda_1}{2}\|f_{i,t}\|_{K_i}^2 \\
    +\lambda_2\sum_{s=1}^{t-1}(f_{i,t}(\x_s^{S_i})-f_{i,t}(\x_t^{S_i}))^2w_{st},\,i=1,2.
    \end{split}
\end{equation}
where $l$ is the number of labeled samples, $\ell$ is the loss function which is convex in its first argument, $f_{i,t}$ is the predictor learned in $i$th feature space and $w_{st}$ is the edge weight which defines a graph over the $T$ samples such as a fully connected graph with Gaussian weights $w_{st}=e^{-\|\x_s-\x_t\|^2/2\sigma^2}$. The last term in $J_{i,t}$ involves the graph edges from $\x_t^{S_i}$ to all previous samples up to time $t$. $\frac{T}{l}$ in the first term of~(\ref{equation:instan risk}) is the empirical estimate of the inverse label probability $1/p_l$, which we assume is given and easily determined based on the rate at which humans can label the data at hand.

The online gradient descent algorithm applied on the instantaneous regularized risk $J_{i,t}$ will derive
\begin{equation}
    \label{equation:general update}
    f_{i,t+1}=\Pi_{\mathcal{H}_{k_i}}\left(f_{i,t}-\tau_t\nabla J_{i,t}(f_{i,t}(\x_t^{S_i}))\right),\,i=1,2,
\end{equation}
where $\tau_t$ is a time-varing step size. Thus even if no label is revealed, we can still update our model $f_{i,t}$ according to (\ref{equation:general update}). Then in round $t>T_1$, the learner can calculate two base predictions based on models $f_{1,t}$ and $f_{2,t}$: $p_{1,t}=f_{1,t}(\psi(\x_t^{S_2}))$ and $p_{2,t}=f_{2,t}(\x_t^{S_2}).$ By ensemble over the two base predictions in each round, our SF$^2$EL is able to follow the best base prediction empirically and theoretically. The initialization process to obtain the relationship mapping $\psi$ and $f_{1,T_1}$ during rounds $1,\ldots,T_1$ is summarized in Algorithm~\ref{alg:Initialize}.

\begin{algorithm}[!t]
	\centering
	\caption{Initialize}
	\label{alg:Initialize}
	\begin{algorithmic}[1]
	\small
		\STATE Initialize $f_{1,1}\in\mathcal{H}_K$ randomly;
		\FOR{$t=1,2,\ldots,T_1$}
	    \STATE Receive $\x_t^{S_1}\in\mathbb{R}^{d_1}$ and predict $p_t=f_{1,t}(\x_t^{S_1})\in\mathbb{R}$;
	    \STATE Receive the target $y_t\in\mathbb{R}$ with small probability $p_i$, and suffer instantaneous risk $J_{i,t}$ according to~(\ref{equation:instan risk});
	    \STATE Update $f_{1,t}$ using~(\ref{equation:general update}) where $\tau_t=1/\sqrt{t}$;
	    \IF{$t>T_1-B$}
	    \STATE Learn $\psi$ by least squares;
	    \ENDIF
        \ENDFOR
    \end{algorithmic}
\end{algorithm}

\subsection{Combining Base Learners}
We propose to do ensemble by combining base learners with weights based on exponential of the cumulative risk~\cite{DBLP:books/daglib/0016248}. The prediction of our method at time $t$ is the weighted average of all the base predictions:
\begin{equation}
\label{equation:average}
\widehat{p}_t=\sum_{i=1}^2\alpha_{i,t}p_{i,t},
\end{equation}
where $\alpha_{i,t}$ is the weight of the $i$th base prediction.
With the previous risk of each base model, we can update the weights of the two base models as follows:
\begin{equation}
\label{equation:update weights_initial}
\alpha_{i,t}=\frac{e^{-\eta \mathcal{J}_{i,t}}}{\sum_{j=1}^2 e^{-\eta \mathcal{J}_{j,t}}},\,i=1,2,
\end{equation}
where $\eta$ is a tuned parameter and $\mathcal{J}_{i,t}$ is the cumulative risk of the $i$th base model until time $t$:
$
\mathcal{J}_{i,t}=\sum_{s=1}^t J_{i,t},\,i=1,2.
$
The risk of our predictor is calculated by
\begin{equation}
    \label{equation:our instan risk}
    J_t=\sum_{i=1}^2\alpha_{i,t}J_{i,t}.
\end{equation}

We can also rewrite (\ref{equation:update weights_initial}) in an incremental way, which can be calculated more efficiently:
\begin{equation}
\label{equation:update weights}
\alpha_{i,t+1}=\frac{\alpha_{i,t}e^{-\eta J_{i,t}}}{\sum_{j=1}^2 \alpha_{j,t}e^{-\eta J_{i,t}}},\,i=1,2.
\end{equation}

The updating rule of the weights shows that if the risk of one of the models on previous round is large, then its weight will decrease in next round, which is reasonable and can derive a good theoretical result shown in Theorem~\ref{theorem:SF2EL}. Thus the procedure of our learning is that we first learn a model $f_{1,T_1}$ using (\ref{equation:general update}) on rounds $1,\ldots,T_1$, during which, we also learn a relationship $\psi$ for $t=T_1-B+1,\ldots,T_1$. Then for $t=T_1+1,\ldots,T_1+T_2$, we learn a model $f_{2,t}$ on each round with new data $\x_t^{S_2}$ from feature space $S_2$:
\begin{equation}
\label{equation:update f2}
f_{2,t+1}=\Pi_{\mathcal{H}_{k_2}}\left(f_{2,t}-\tau_t\nabla J_{2,t}(f_{2,t}(\x_t^{S_2}))\right)
\end{equation}
and keep updating $f_{1,t}$ on the recovered data $\psi(\x_t^{S_2})$:
\begin{equation}
\label{equation:update f1}
f_{1,t+1}=\Pi_{\mathcal{H}_{k_1}}\left(f_{1,t}-\tau_t\nabla J_{1,t}(f_{1,t}(\psi(\x_t^{S_2})))\right),
\end{equation}
where $\tau_t$ is a varied step size.
Then we combine the predictions of the two models by weights calculated in (\ref{equation:update weights}).


In order to compute~(\ref{equation:update f2}) and (\ref{equation:update f1}), we first need to compute their gradients $\nabla J_{i,t}(f_{i,t}(\x_t^{S_i})),\,i=1,2$. We express the functions in $i$th feature space $f_{i,1},\ldots,f_{i,t},\,i=1,2$ using a common set of representers $\x_1^{S_i},\ldots,\x_t^{S_i},\,i=1,2$, i.e.,
\begin{equation}
\label{equation:kernel expansion}
f_{i,t}=\sum_{s=1}^{t-1}\beta_{i,s}^{(t)}K_i(\x_s^{S_i},\cdot),\,i=1,2.
\end{equation}
To obtain $f_{i,t+1},\,i=1,2$, we need to calculate the coefficients $\beta_{i,1}^{(t+1)},\ldots,\beta_{i,t}^{(t+1)}$. We follow the kernel online semi-supervised learning approach~\cite{DBLP:conf/pkdd/GoldbergLZ08} to update our coefficients by writing the gradient $\nabla J_{i,t}(f_{i,t}(\x_t^{S_i})),\,i=1,2$ as
\begin{equation}
    \label{equation:kernel general gradient}
    \resizebox{.89\linewidth}{!}{$
    \begin{split}
    &\frac{T}{l}\delta(y_t)\ell'(f_{i,t}(\x_t^{S_i}),y_t)K_i(\x_t^{S_i},\cdot)+\lambda_1f_{i,t}\\
    +&2\lambda_2\sum_{s=1}^{t-1}(f_{i,t}(\x_s^{S_i})-f_{i,t}(\x_t^{S_i}))w_{st}(K_i(\x_s^{S_i},\cdot)-K_i(\x_t^{S_i},\cdot)),
    \end{split}$}
\end{equation} 
in which we compute the derivative according to the reproducing property of RKHS, i.e., \[\partial f_{i,t}(\x_t^{S_i})/\partial f_{i,t}=\partial \langle f_{i,t},K_i(\x_t^{S_i},\cdot)\rangle/\partial f_{i,t}=K_i(\x_t^{S_i},\cdot),\]
where $i=1,2.$ $\ell'$ is the (sub)gradient of the loss function $\ell$ with respect to $f_{i,t}(\x_t^{S_i}),\,i=1,2$. Putting (\ref{equation:kernel general gradient}) back to (\ref{equation:update f2}) or (\ref{equation:update f1}), and replace $f_{i,t}$ with its kernel expansion (\ref{equation:kernel expansion}), we can obtain the coefficients for $f_{i,t+1}$ as follows:
\begin{equation}
\label{equation:coefficients update1}
\beta_{i,s}^{(t+1)}=(1-\tau_t\lambda_1)\beta_{i,s}^{(t)}-2\tau_t\lambda_2(f_{i,t}(\x_s^{S_i})-f_{i,t}(\x_t^{S_i}))w_{st},
\end{equation}
where $i=1,2$ and $s=1,\ldots,t-1$, and
\begin{equation}
\label{equation:coefficients update2}
\begin{split}
&\beta_{i,t}^{(t+1)}=2\tau_t\lambda_2\sum_{s=1}^{t-1}(f_{i,t}(\x_s^{S_i})-f_{i,t}(\x_t^{S_i}))w_{st}\\
&-\tau_t\frac{T}{l}\delta(y_t)\ell'(f_{i,t}(\x_t^{S_i}),y_t),\,i=1,2.
\end{split}
\end{equation}

\subsection{Buffering}
\label{section:buffering}
As can be seen from (\ref{equation:coefficients update1}) and (\ref{equation:coefficients update2}), when updating the model, we need to store each observed sample and calculate the weights $w_{st}$ between the new incoming sample and all the other observed ones. These operations will bring huge burdens on computation and storage. To alleviate this problem, we do not store all the observed samples. Instead, we use a buffer to store a small part of them, which we call \emph{buffering}.

We denote by $\mathcal{B}$ the buffer and let its size be $b$. In order to make the samples in buffer more representative, it is better to make each sample in the buffer sampled by equal quality. Therefore, we exploit the \emph{reservoir sampling} technique~\cite{DBLP:journals/toms/Vitter85} to achieve this goal which enables us to use a fixed size buffer to represent all the received samples. Specifically, when receiving a sample $\x_t^{S_i}$, we will directly add it to the buffer if the buffer size $b>t$. Otherwise, with probability $b/t$, we update the buffer $\mathcal{B}$ by randomly replacing one sample in $\mathcal{B}$ with $\x_t^{S_i}$. The key property of reservoir sampling is that \emph{samples in the buffer are provably sampled from the original dataset uniformly}. Then the instantaneous risk will be approximated by
\begin{equation}
    \label{equation:instan risk with buffer}
    \resizebox{.85\linewidth}{!}{$
    \begin{split}
    J_{i,t}(f_{i,t}(\x_t^{S_i}))=\frac{T}{l}\delta(y_t)\ell(f_{i,t}(\x_t^{S_i}),y_t)+\frac{\lambda_1}{2}\|f_{i,t}\|_{K_i}^2 \\
    +\lambda_2\frac{t-1}{b}\sum_{s\in\mathcal{B}}(f_{i,t}(\x_s^{S_i})-f_{i,t}(\x_t^{S_i}))^2w_{st},\,i=1,2,
    \end{split}$}
\end{equation}
where the scaling factor $\frac{t-1}{b}$ keeps the magnitude of the manifold regularizer comparable to that of the unbuffered one. Accordingly, the predictor will become
\begin{equation}
\label{equation:kernel expansion with buffer}
f_{i,t}=\sum_{s\in\mathcal{B}}\beta_{i,s}^{(t)}K_i(\x_s^{S_i},\cdot),\,i=1,2.
\end{equation}

If the buffer size $b>t$, we will update the coefficients by (\ref{equation:coefficients update1}) and (\ref{equation:coefficients update2}) directly. Otherwise, if the new incoming sample replaces some sample in the buffer, there will be two steps to update our predictor. The first step is to update $f_{i,t}$ to an intermediate function $f'$ represented by $b+1$ elements including the old buffer and the new observed sample $\x_t^{S_i}$ as follows.
\begin{equation}
    f'=\sum_{s\in\mathcal{B}}\beta_{i,s}'K_i(\x_s^{S_i},\cdot)+\beta_{i,t}'K_i(\x_t^{S_i},\cdot),
\end{equation}
where
\begin{equation}
    \label{equation:coefficients update1 with buffer}
    \beta_{i,s}'=(1-\tau_t\lambda_1)\beta_{i,s}^{(t)}-2\tau_t\lambda_2(f_{i,t}(\x_s^{S_i})-f_{i,t}(\x_t^{S_i}))w_{st},
\end{equation}
in which $i=1,2$ and $s\in\mathcal{B}$, and
\begin{equation}
    \label{equation:coefficients update2 with buffer}
    \begin{split}
    &\beta_{i,t}'=2\tau_t\lambda_2\frac{t-1}{b}\sum_{s\in\mathcal{B}}(f_{i,t}(\x_s^{S_i})-f_{i,t}(\x_t^{S_i}))w_{st}\\
    &-\tau_t\frac{T}{l}\delta(y_t)\ell'(f_{i,t}(\x_t^{S_i}),y_t),\,i=1,2.
    \end{split}
\end{equation}
The second step is to use the newest sample $\x_t^{S_i}$ to replace the sample selected by reservoir sampling, say $\x_s^{S_i}$ and obtain $f_{i,t+1}$ which uses $b$ base representers by approximating $f'$ which uses $b+1$ base representers: 
\begin{equation}
\label{equation:matching pursuit}
\begin{split}
    \min_{\beta_i^{(t+1)}}~&\|f'-f_{i,t+1}\| \\
    \text{s.t.}~~~&\,f_{i,t+1}=\sum_{s\in\mathcal{B}}\beta_{i,s}^{(t+1)}K_i(\x_s^{S_i},\cdot),\,i=1,2.
\end{split}
\end{equation}
This can be intuitively regarded as spreading the replaced weighted contribution $\beta_{i,s}'K_i(\x_{s}^{S_i},\cdot)$ to the remaining samples including the newly added $\beta_{i,t}'K_i(\x_{t}^{S_i},\cdot)$ in the buffer. The optimal $\beta_i^{(t+1)}$ in (\ref{equation:matching pursuit}) can be efficiently found by \emph{matching pursuit}~\cite{DBLP:journals/ml/VincentB02}. 

\begin{algorithm}[!t]
	\centering
	\caption{SF$^2$EL}
	\label{alg:SF2EL}
	\begin{algorithmic}[1]
	\small
		\STATE Initialize $\psi$ and $f_{1,T_1}$ during $1,\ldots,T_1$ using Algorithm~\ref{alg:Initialize};
	    \STATE $\alpha_{1,T_1}=\alpha_{2,T_1}=\frac{1}{2}$;
	    \STATE Initialize $f_{2,T_1+1}$ randomly and $f_{1,T_1+1}$ by $f_{1,T_1}$;
	    \FOR{$t=T_1+1,T_1+2,\ldots,T_1+T_2$}
	    \STATE Receive $\x_t^{S_2}\in\mathbb{R}^{S_2}$;
	    \STATE Predict $p_{1,t}=f_{1,t}(\psi(\x_t^{S_2}))\;\text{and}\;p_{2,t}=f_{2,t}(\x_t^{S_2})$;
	    \STATE Predict $\widehat{p}_t\in\mathbb{R}$ using~(\ref{equation:average});
	    \STATE Receive the target $y_t\in\mathbb{R}$ with small probability $p_i$, and suffer instantaneous risk $J_t$ according to~(\ref{equation:our instan risk});
	    \STATE Update base predictions' weights using~(\ref{equation:update weights});
	    \STATE Update $f_{1,t}$ and $f_{2,t}$ using~(\ref{equation:update f1}) and (\ref{equation:update f2}) respectively with \emph{buffering} strategy in Section~\ref{section:buffering} where $\tau_t=1/\sqrt{t-T_1}$.
	    \ENDFOR
	\end{algorithmic}
\end{algorithm}

If the new incoming sample does not replace the sample in the buffer, $f_{i,t+1}$ will still consist of the representers from the unchanged buffer. Then only the coefficients of the representers from the buffer will be updated as follows.
\begin{equation}
    \label{equation:coefficients update1 with buffer2}
    \beta_{i,s}^{(t+1)}=(1-\tau_t\lambda_1)\beta_{i,s}^{(t)}-2\tau_t\lambda_2(f_{i,t}(\x_s^{S_i})-f_{i,t}(\x_t^{S_i})),
\end{equation}
where $i=1,2$ and $s\in\mathcal{B}$. 

Algorithm~\ref{alg:SF2EL} summarizes our SF$^2$EL.


\subsection{Analysis}
In this section, we borrow \emph{regret} from online learning to measure the performance of SF$^2$EL. Specifically, we give a risk bound which demonstrates that the performance will be improved with the assistance of the old feature space. We define that $\mathcal{J}^{S_1}$ and $\mathcal{J}^{S_2}$ are two cumulative risks suffered by base models on rounds $T_1+1,\ldots,T_1+T_2$,
$
\mathcal{J}^{S_1}=\sum_{t=T_1+1}^{T_1+T_2}J_{1,t}$, $
\mathcal{J}^{S_2}=\sum_{t=T_1+1}^{T_1+T_2}J_{2,t},
$  
and $\mathcal{J}^{S_{12}}$ is the cumulative risk suffered by our method according to the definition of our predictor's risk in (\ref{equation:our instan risk}): 
$
\mathcal{J}^{S_{12}}=\sum_{t=T_1+1}^{T_1+T_2} J_t.
$
Then we have (proof is deferred to supplementary file):
\begin{theorem}
\label{theorem:SF2EL}
Assume that the risk function $J_t$ takes value in [0,1]. For all $T_2>1$ and for all $y_t\in\mathcal{Y}$ with $t=T_1+1,\ldots,T_1+T_2$, $\mathcal{J}^{S_{12}}$ with parameter $\eta=\sqrt{\ln2/T_2}$ satisfies
\begin{equation}
\label{equation:SF2EL}
\mathcal{J}^{S_{12}}\leq \min(\mathcal{J}^{S_1},\mathcal{J}^{S_2})+\sqrt{T_2\ln2}.
\end{equation}
\end{theorem}
\begin{remark}
This theorem implies that the cumulative risk $\mathcal{J}^{S_{12}}$ of Algorithm~\ref{alg:SF2EL} over rounds $T_1+1,\ldots,T_1+T_2$ is comparable to the minimum of $\mathcal{J}^{S_1}$ and $\mathcal{J}^{S_2}$. Furthermore, we define $C=\sqrt{T_2\ln2}$. If $\mathcal{J}^{S_2}-\mathcal{J}^{S_1}>C$, it is easy to verify that $\mathcal{J}^{S_{12}}$ is smaller than $\mathcal{J}^{S_2}$. In summary, on rounds $T_1+1,\ldots,T_1+T_2$, when $f_{1,t}$ is better than $f_{2,t}$ to certain degree, the model with assistance from $S_1$ is better than that without assistance.
\end{remark}

\vspace{-0.1cm}
Furthermore, we prove that larger buffer can bring better performance by leveraging our buffering strategy. Concretely, let $R_t$ be the last term of the objective~\eqref{equation:instan risk}, which is formed by all the observed samples till the current iteration. Denote by $\hat{R}_t$ the approximated version formed by the observed samples in the buffer. Then we have:
\begin{theorem}
	\label{thm:unbiased}
	With the reservoir sampling mechanism, the approximated objective is an unbiased estimation of objective formed by the original data, namely, $\mathbb{E}[R_t]=\mathbb{E}[\hat{R}_t]$.
\end{theorem}

\begin{remark}
Theorem~\ref{thm:unbiased} demonstrates the rationality of the reservoir sampling mechanism in buffering. The objective formed by the observed samples in the buffer is provably unbiased to that formed by all the observed samples. Furthermore, the variance of the approximated objective will decrease with more observed samples in a larger buffer, leading to a more accurate approximation, which suggests us to make the best of the buffer storage to store previous observed samples. Since various devices have different storage budgets and even the same device will provide different available storages, we can fit our method for different storages to maximize the performance by taking full advantage of the budget. This proof can be found in supplementary file.
\end{remark}

\section{Experiments}
\label{section:experiment}
In this section, we conduct experiments in different scenarios to validate the three claims presented in Introduction.

\subsection{Compared Methods}
We compare our SF$^2$EL to $7$ baseline methods:
\begin{itemize}
\item NOGD: (Naive Online Gradient Descent): mentioned in Preliminary, where once the feature space changes, the online gradient descent algorithm will be invoked from scratch. 
\item uROGD (updating Recovered Online Gradient Descent): utilizes the model learned from feature space $S_1$ by online gradient descent to do predictions on the recovered data and keeps updating with the recovered data.
\item fROGD (fixed Recovered Online Gradient Descent): also utilizes the model learned from $S_1$ to do predictions on the recovered data like uROGD but keeps fixed.
\item NOGD+MR: NOGD boosted by manifold regularization (MR).
\item uROGD+MR: uROGD boosted by MR.
\item fROGD+MR: fROGD boosted by MR.
\item FESL-Variant: FESL~\cite{DBLP:conf/nips/Hou0Z17} cannot be directly applied in our setting. For fair comparison, we modify the original FESL to a non-linear version which only updates on the rounds when there is a label revealed. FESL-Variant is actually the SF$^2$EL without MR.
\end{itemize}
Note that NOGD, uROGD, fROGD and FESL-Variant do not update on rounds when no label is revealed while NOGD+MR, uROGD+MR, fROGD+MR and our SF$^2$EL keep updating on every round. We want to emphasize that it is \emph{sufficient} to validate the effectiveness of our framework by merely comparing our method to these baselines mentioned above in the scenario of FESL since our goal is: (1) our model can be comparable to these base models, (2) the manifold regularization is useful and (3) our method can fit for the storage budget to maximize its performance. With the manifold regularization and buffering strategy, other feature evolvable learning methods based on the online learning technique can also adapt to our framework similarly. 

\begin{figure*}[!t]

\centering
\small
\begin{minipage}{0.19\linewidth}\centering
    \includegraphics[width=1\textwidth]{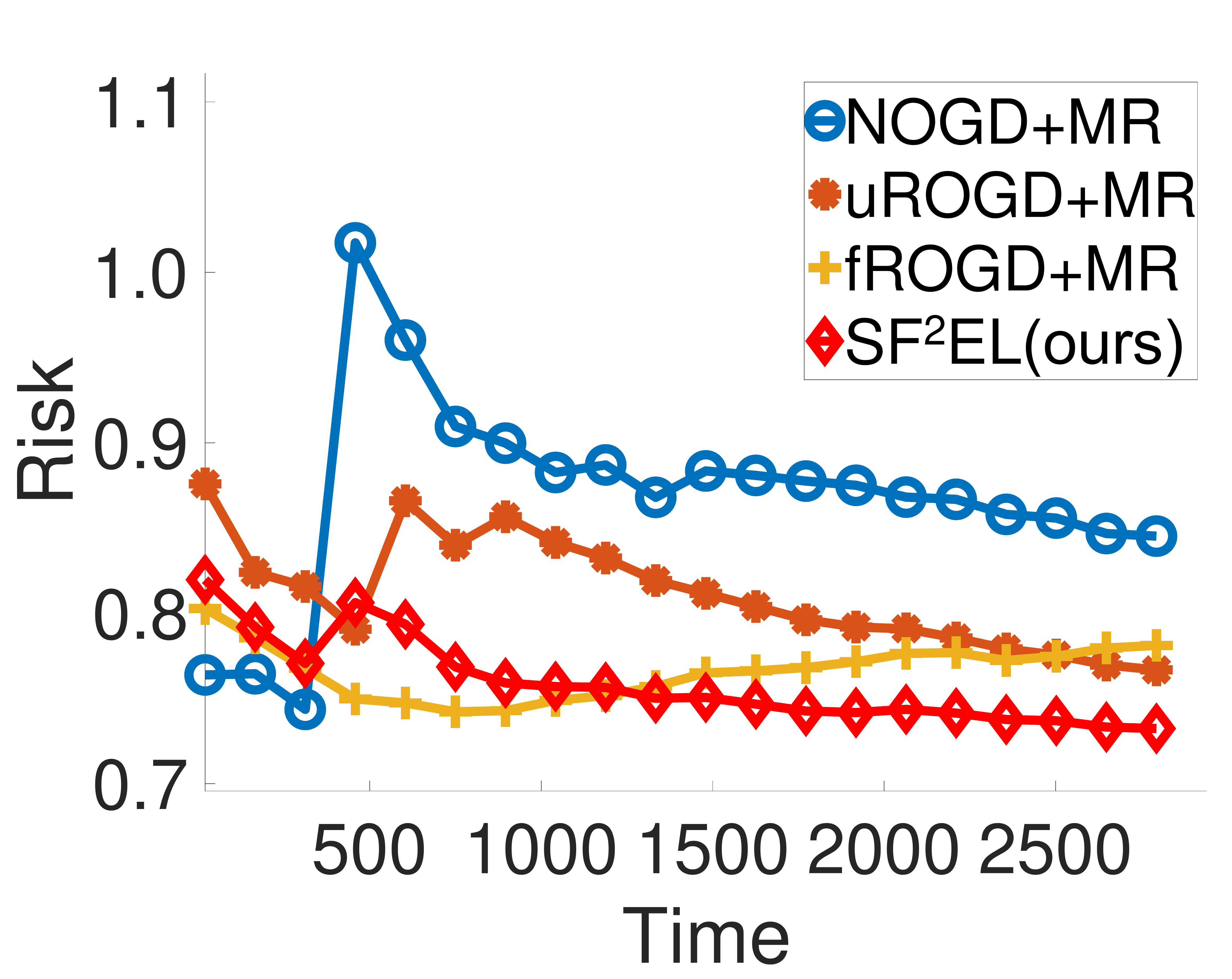}\\
    \mbox{\scriptsize\quad\quad(a) \emph{Credit-a}}
\end{minipage}
\begin{minipage}{0.19\linewidth}\centering
    \includegraphics[width=1\textwidth]{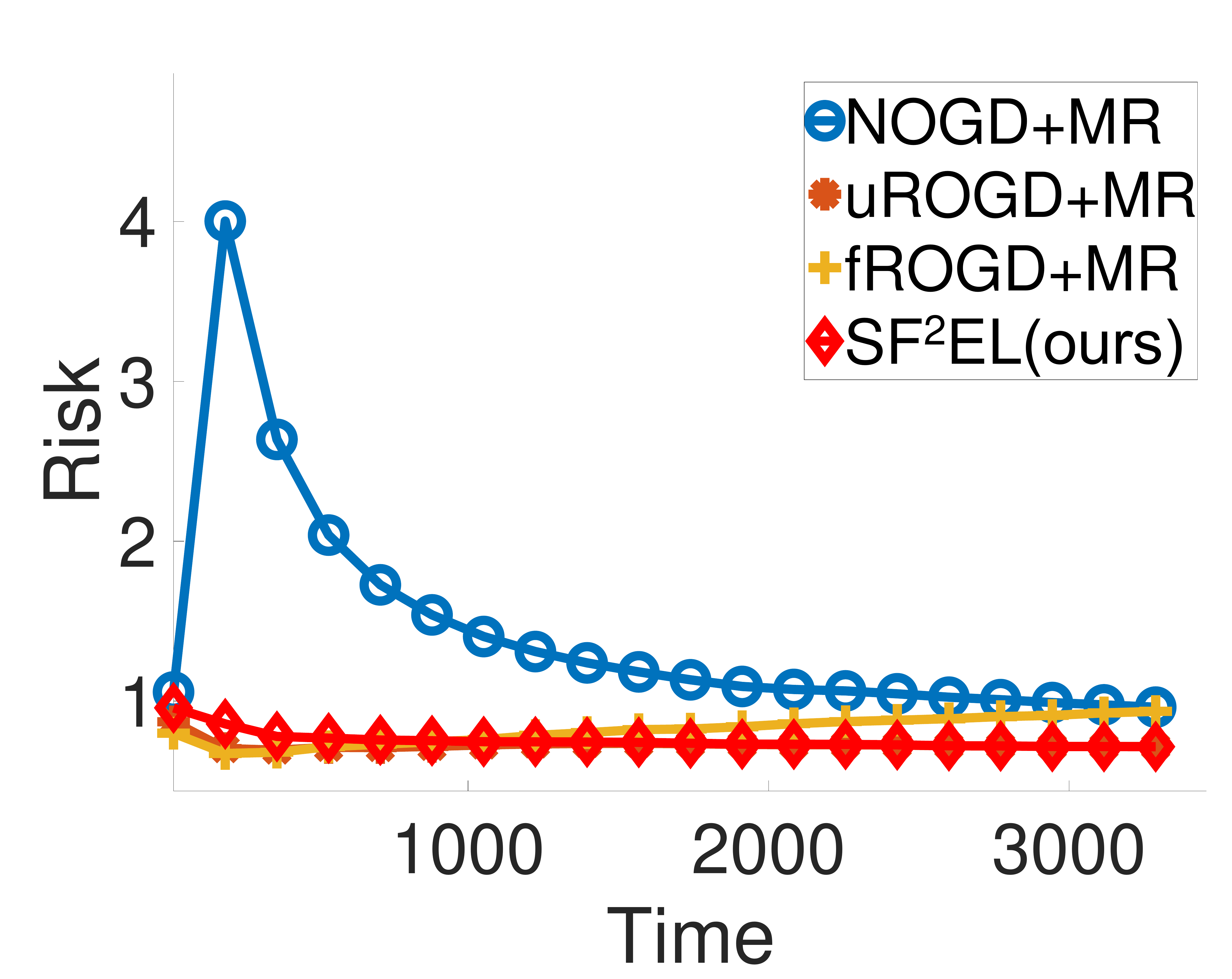}\\
    \mbox{\scriptsize\quad\quad(b) \emph{Diabetes}}
\end{minipage}
\begin{minipage}{0.19\linewidth}\centering
    \includegraphics[width=1\textwidth]{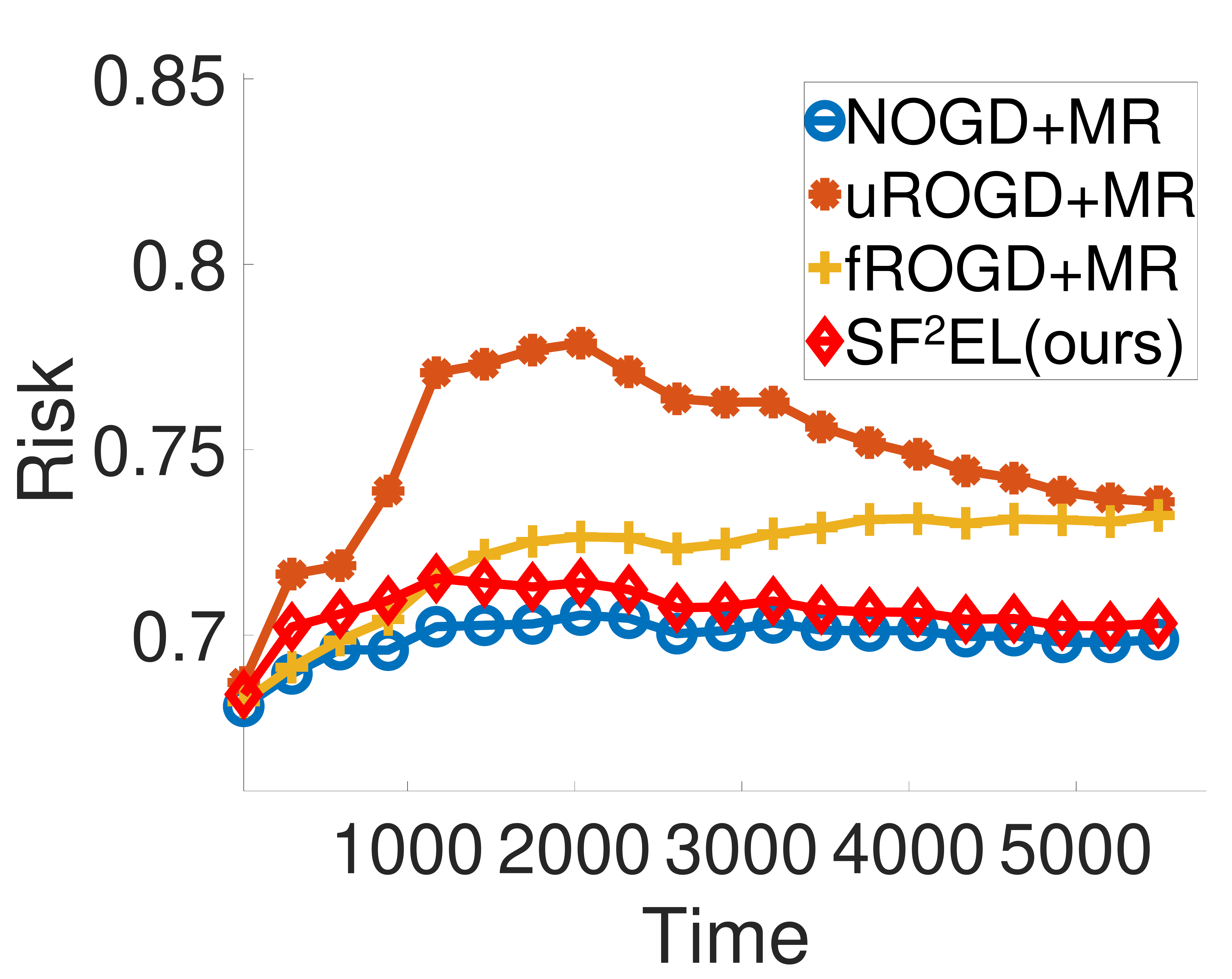}\\
    \mbox{\scriptsize\quad\quad(c) \emph{Svmguide3}}
\end{minipage}
\begin{minipage}{0.19\linewidth}\centering
    \includegraphics[width=1\textwidth]{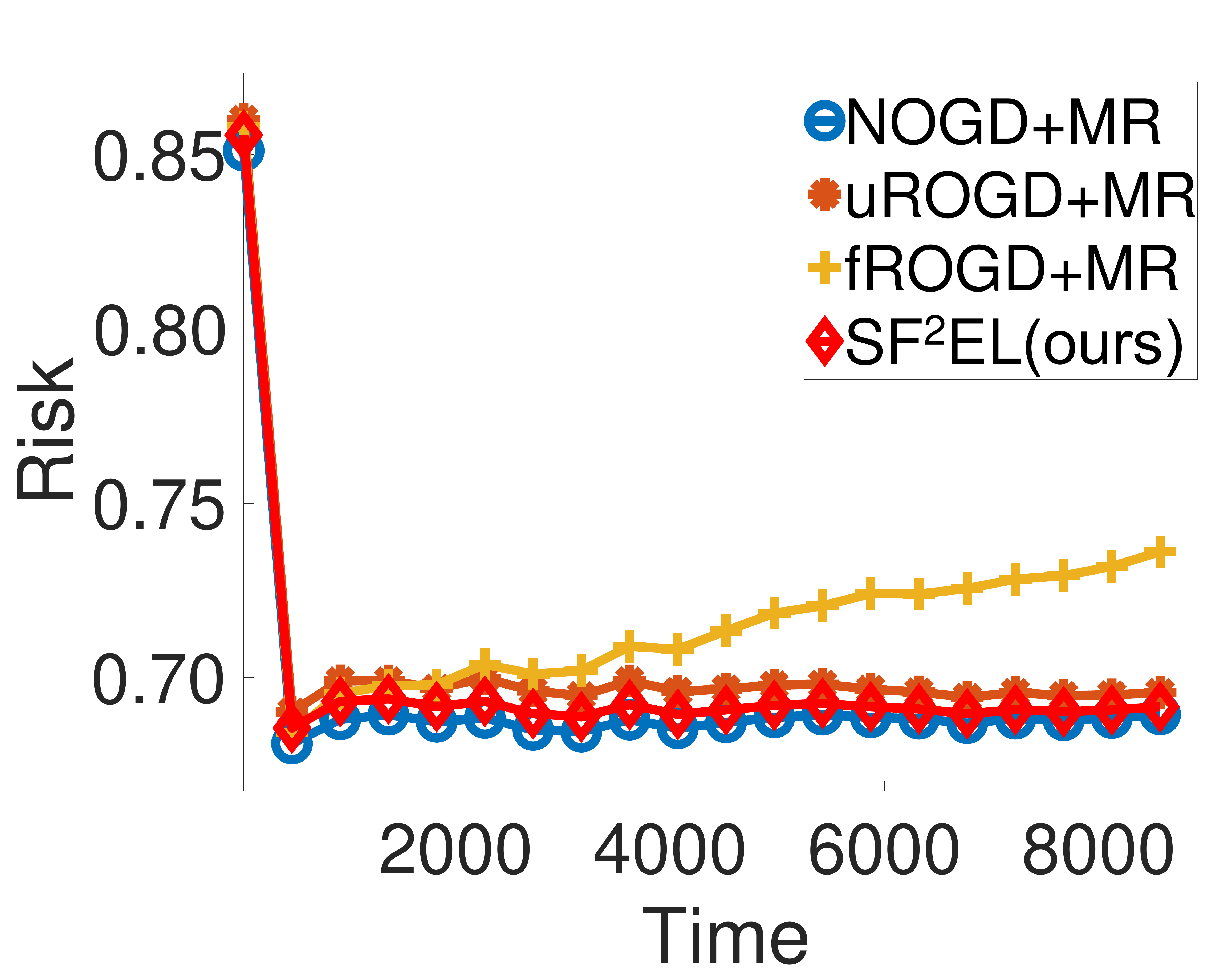}\\
    \mbox{\scriptsize\quad\quad(d) \emph{Swiss}}
\end{minipage}
\begin{minipage}{0.19\linewidth}\centering
    \includegraphics[width=1\textwidth]{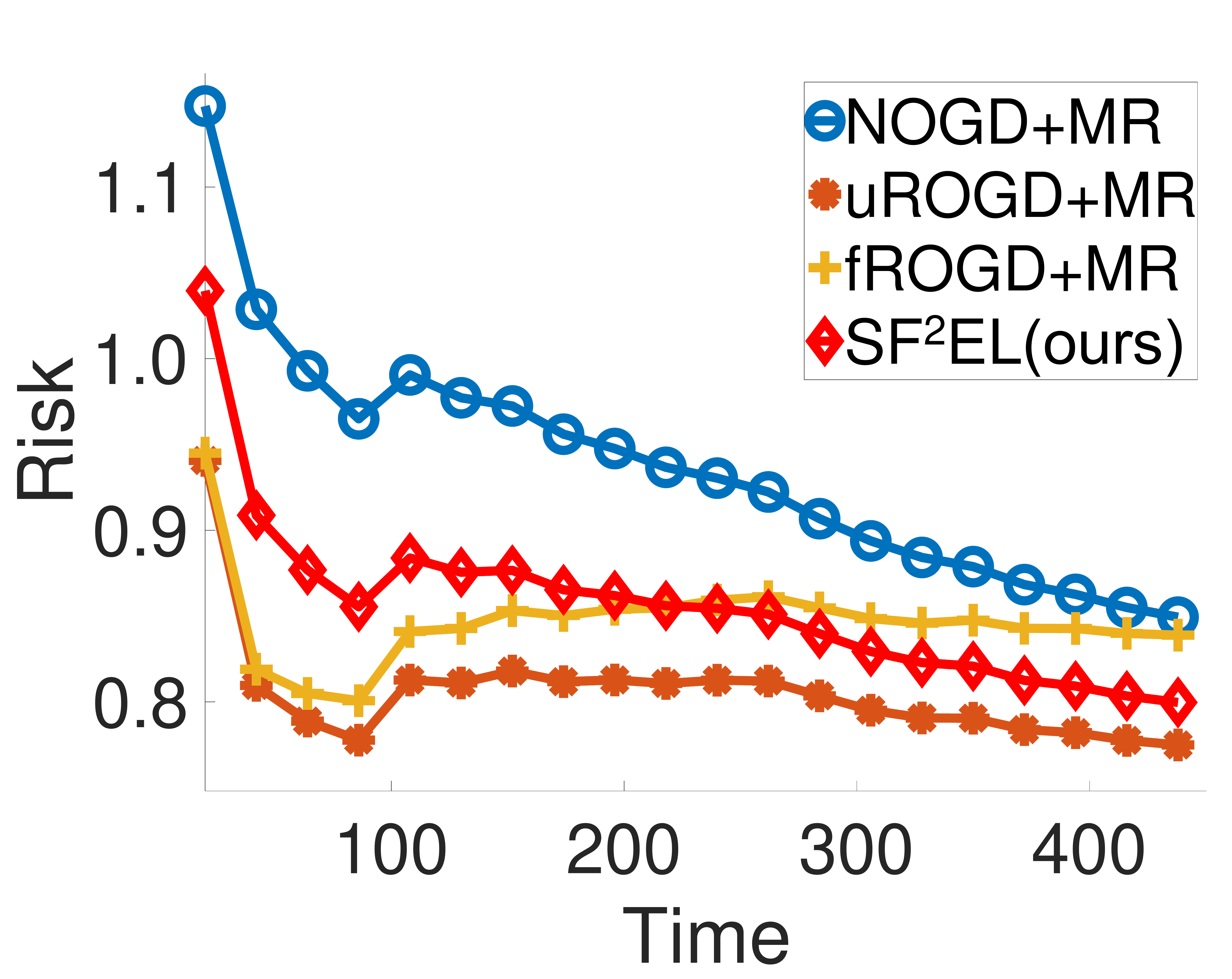}\\
    \mbox{\scriptsize\quad\quad(e) \emph{RFID}}
\end{minipage}
\vspace{-0.4em}
\caption{\small{The trend of risk with NOGD+MR, uROGD+MR, fROGD+MR and SF$^2$EL. We only compare SF$^2$EL with these baselines because those without MR cannot calculate a risk when there is no label. The smaller the cumulative risk is, the better. All the average cumulative risk at any time of our method is comparable to the best baselines. Note that our goal is to be comparable to the best baseline and is not necessary to be better than them. fROGD+MR’s risk sometimes increases because it does not update itself.}}
\label{figure:performance comparisons}
\vspace{-0.2cm}
\end{figure*}

\begin{table*}[!t]
\renewcommand\arraystretch{1.4}
\caption{\small Accuracy (mean$\pm$std) comparisons between baselines and SF$^2$EL when buffer size is $60$. ``+MR" means the baselines are boosted by manifold regularization(MR). Better result in each grid is marked by $\bullet$. The best one among all the methods is bold. Note that our goal is to be comparable to the best baseline and is not necessary to be better than them.}
    \label{table:performance comparisons}
    \centering
    \small
    \setlength\tabcolsep{9pt}
        \begin{tabular}{l|c|c|c|c|c|c|c}
\hline
\makecell[c]{Dataset} & \makecell[c]{Credit-a}  & \makecell[c]{Diabetes} & \makecell[c]{Svmguide3} & \makecell[c]{Swiss} & \makecell[c]{RFID} & \makecell[c]{HTRU\_2} & \makecell[c]{magic04}\\  
\hline
NOGD & .690$\pm$.051 & .643$\pm$.029 & .657$\pm$.036 & .711$\pm$.044 & .687$\pm$.042 & .885$\pm$.022 & .580$\pm$.120\\
NOGD+MR & $\bullet$.706$\pm$.049 & $\bullet$.672$\pm$.019 & $\bullet$.668$\pm$.037 & $\bullet$.807$\pm$.026 & $\bullet$.688$\pm$.042 & $\bullet$ .907$\pm$.001 & $\bullet$.616$\pm$.058\\
\hline
uROGD & .740$\pm$.038 & .658$\pm$.021 & .675$\pm$.045 & .694$\pm$.071 & .571$\pm$.036 & .943$\pm$.014 & .550$\pm$.172\\
uROGD+MR & $\bullet$.760$\pm$.034 & $\bullet$.678$\pm$.015 & $\bullet$.680$\pm$.048 & $\bullet$.824$\pm$.050 & $\bullet$.572$\pm$.036 & $\bullet$\textbf{.944$\pm$.010} & $\bullet$.603$\pm$.079\\
\hline
fROGD & .672$\pm$.087 & .633$\pm$.056 & .648$\pm$.035 & .702$\pm$.073 & .560$\pm$.045 & .757$\pm$.179 & .550$\pm$.172\\
fROGD+MR & $\bullet$.697$\pm$.079 & $\bullet$.654$\pm$.041 & $\bullet$.659$\pm$.037 &$\bullet$ .811$\pm$.067 & $\bullet$.561$\pm$.045 & $\bullet$.943$\pm$.020 & $\bullet$\textbf{.649$\pm$.001}\\
\hline
FESL-Variant & .759$\pm$.028 & .666$\pm$.018 & .686$\pm$.039 & .855$\pm$.034 & .688$\pm$.040 & .885$\pm$.022 & .556$\pm$.162\\
SF$^2$EL (ours) & $\bullet$\textbf{.768$\pm$.029} & $\bullet$\textbf{.685$\pm$.011} & $\bullet$\textbf{.694$\pm$.040} & $\bullet$\textbf{.939$\pm$.020} & $\bullet$\textbf{.690$\pm$.040} & $\bullet$.912$\pm$.008 & $\bullet$\textbf{.649$\pm$.001}\\
\hline
        \end{tabular}
        
\end{table*}

\subsection{Evaluation and Parameter Setting}
We evaluate the empirical performances of the proposed approaches on classification task on rounds $T_1+1,\ldots,T_1+T_2$. We assume all the labels can be obtained in hindsight. Thus the accuracy is calculated on all rounds. Besides, to verify that Theorem~\ref{theorem:SF2EL} is reasonable, we present the trend of average cumulative risk. Concretely, at each time $t'$, the risk $\bar{J}_{i,t'}$ of every method is the average of the cumulative risk over $1,\ldots,t'$, namely $ \bar{J}_{i,t'}=(1/t')\sum\nolimits_{t=1}^{t'}J_{i,t}$. The probability of labeled data $p_l$ is set as $0.3$. We also conduct experiments on other different $p_l$ and our SF$^2$EL also works well. The performances of all approaches are obtained by average results over $10$ independent runs. 

   \begin{figure}
    \begin{minipage}[t]{\linewidth}
       \centering
  \begin{subfigure}[]{
    \begin{minipage}[b]{0.42\textwidth}
    \label{figure:swiss_MR}
     \includegraphics[width=\textwidth]{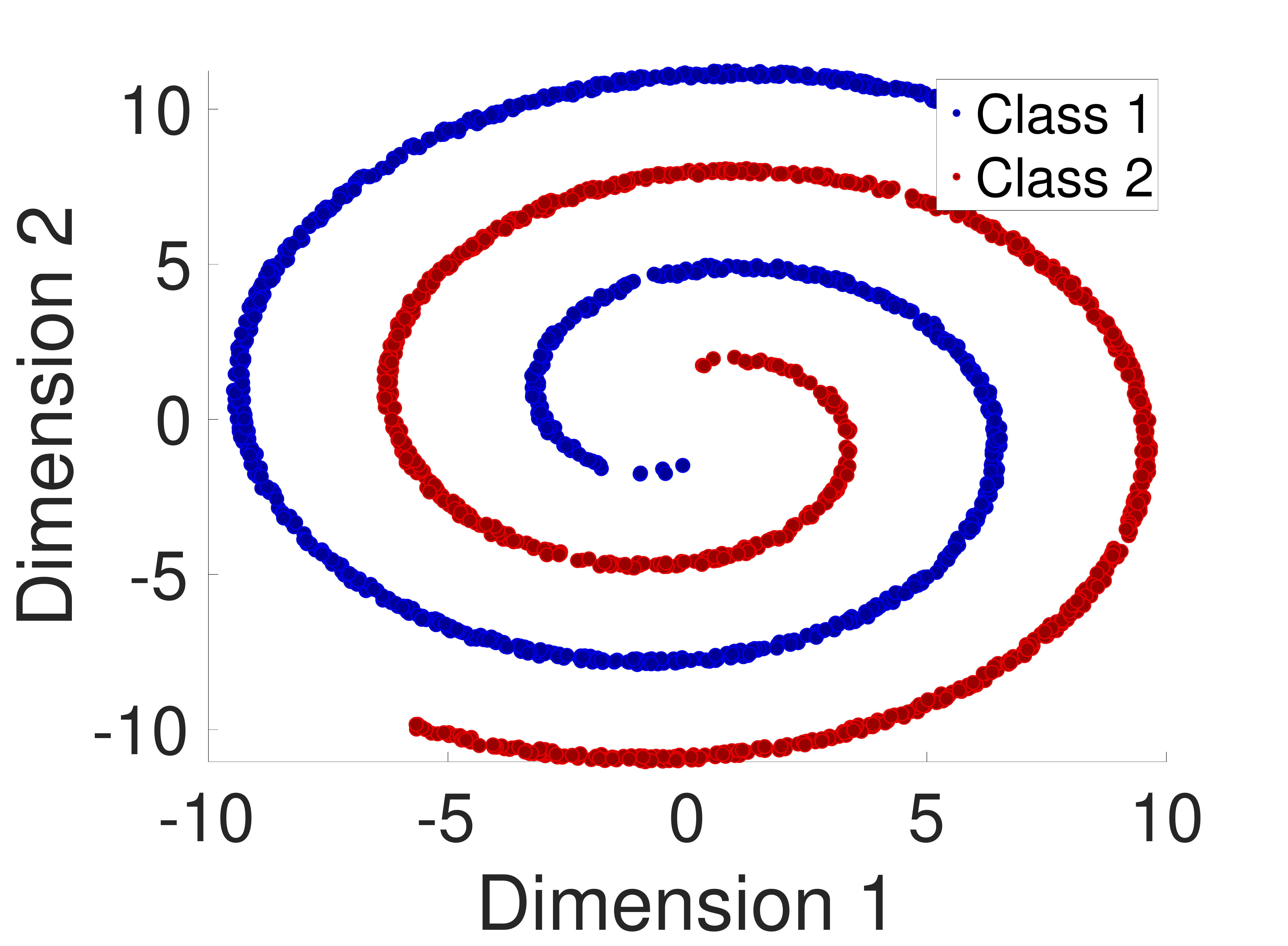}
     \end{minipage}}
  \end{subfigure}
  \begin{subfigure}[]{
    \begin{minipage}[b]{0.42\textwidth}
     \label{figure:buffer_size_impact}
     \includegraphics[width=\textwidth]{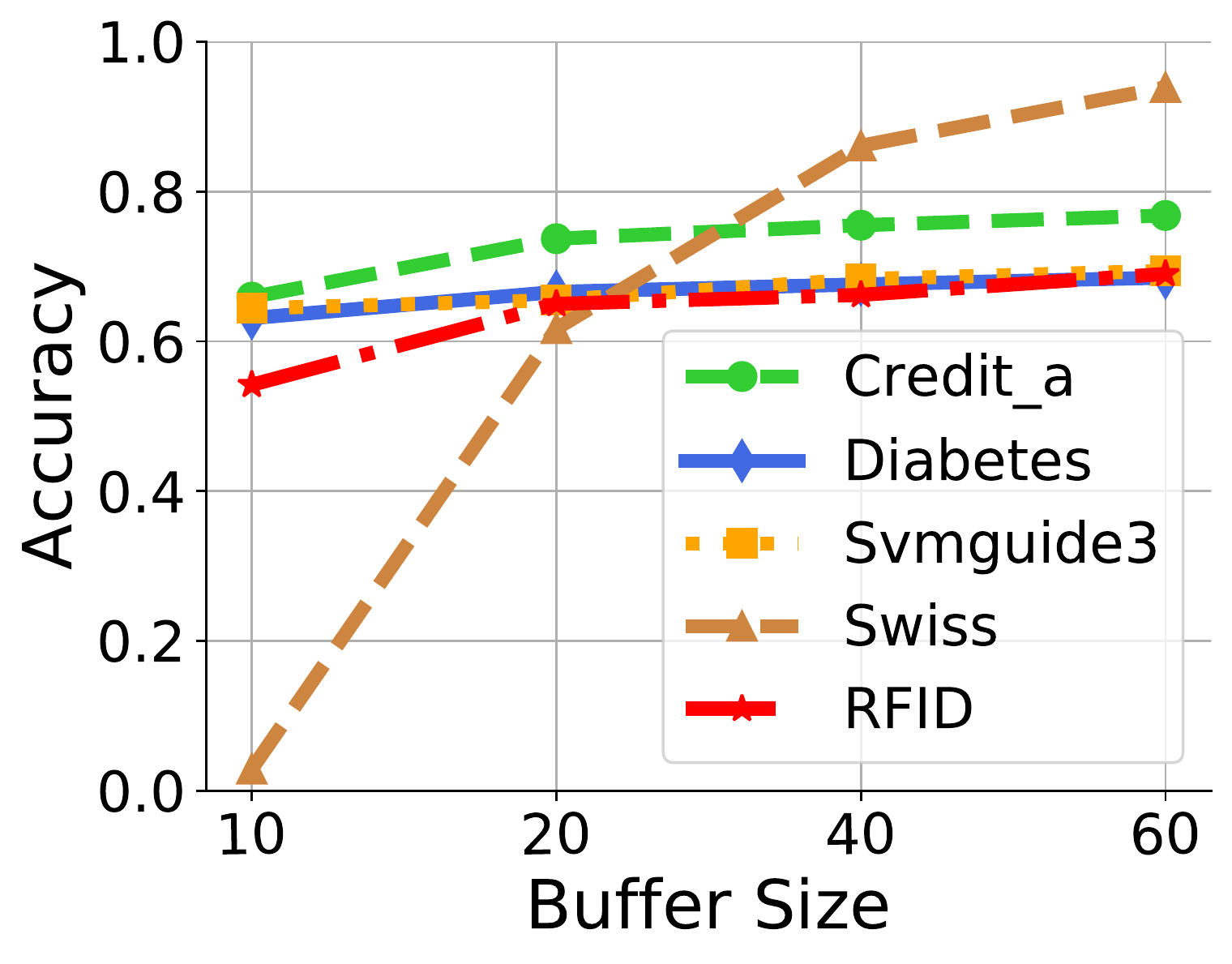}
     \end{minipage}}
  \end{subfigure}
  \vspace{-0.4cm}
  \caption{\small (a) is the manifold of Swiss dataset.  (b) exhibits the impact of buffer size on accuracy.}
    \label{fig}
  \end{minipage}  
    \vspace{-3.5mm}
\end{figure}

\subsection{Datasets}
We conduct our experiments on $7$ datasets from different domains including \emph{economy} and \emph{biology}, etc.\footnote{The details of the datasets including their sources, descriptions and dimensions can be found in supplementary file.} Note that in FESL $30$ datasets are used. However, over $20$ of them are the texting datasets which do not satisfy the manifold characteristic. The datasets used in our paper all satisfy the manifold characteristic and the Swiss dataset (like a swiss roll) is the perfect one. Swiss is a synthetic dataset containing $2000$ samples and is generated by two twisted spiral datasets. As Swiss has only two dimensions, it is convenient for us to observe its manifold characteristic. As can be seen from Figure~\ref{figure:swiss_MR}, Swiss satisfies a pretty nice manifold property. Other datasets used in our paper also have such property but as a matter of high dimension, we only use Swiss as an example. To generate synthetic data of feature space $S_2$, we artificially map the original datasets by random matrices. Then we have data both from feature space $S_1$ and $S_2$. Since the original data are in batch mode, we manually make them come sequentially. In this way, synthetic data are completely generated. As for the real dataset, we use ``RFID'' dataset provided by FESL which satisfies all the assumptions in Preliminary. ``HTRU\_2'' and ``magic04'' are two large-scale datasets which contain $17898$ and $19020$ instances respectively and we only provide their accuracy results in Table~\ref{table:performance comparisons} due to page limitation. Other results on these two datasets can be found in the supplementary file.

\subsection{Results}
We have three claims mentioned in Introduction. The first is that our method can always follow the best baseline at any time and thus achieve the fundamental goal of feature evolvable learning: always keeps the performance at a good level. The second is that manifold regularization brings better performance when there are only a few labels. The last one is that larger buffer will bring better performance and thus our method can fit for different storages by taking full advantage of the budget. In the following, we show the experimental results that validate these three claims.

\subsubsection{Following the Best Baseline}
Figure~\ref{figure:performance comparisons} shows the trend of risk of our method and the baselines boosted by manifold regularization. We only compare SF$^2$EL with baselines boosted by MR because those without MR cannot calculate a risk when there is no label. The smaller the cumulative risk is, the better. fROGD+MR's risk sometimes increases because it does not update itself.
Note that our goal is let our model be comparable to the best baseline yet is not necessary to be better than them. We can see that our method's risk is always comparable with the best baseline which validates Theorem~\ref{theorem:SF2EL}. And surprisingly, as can be seen from Table~\ref{table:performance comparisons}, our method's accuracy results on classification tasks almost outperform those of the baseline methods ($6$ out of $7$). 

\subsubsection{MR Brings Better Performance}
In Table~\ref{table:performance comparisons}, we can see that MR makes NOGD, fROGD and uROGD better, and our method also benefits from it. Specifically, our SF$^2$EL is based on the ensemble of uROGD+MR and NOGD+MR, which makes it the best in all datasets. FESL-Variant is based on NOGD and uROGD. Although it is better than NOGD and uROGD, it is worse than our SF$^2$EL.

\begin{table}[!t]
\renewcommand\arraystretch{1.7}
\caption{\small Accuracy (mean$\pm$std) comparisons with different buffer sizes. The best ones among all the buffers are bold. We can find that larger buffer brings better performance.}
    \label{table:larger buffer better performance}
    \centering
    \small
    \setlength\tabcolsep{1pt}
        \begin{tabular}{c|c|c|c|c|c}
\hline
\makecell[c]{Buffer} & \makecell[c]{Credit-a} & \makecell[c]{Diabetes} & \makecell[c]{Svmguide3} & \makecell[c]{Swiss} & \makecell[c]{RFID}\\  
\hline
10 & .659$\pm$.052 & .631$\pm$.066 & .644$\pm$.103 & .290$\pm$.070 & .542$\pm$.056\\
\hline
20 & .737$\pm$.036 & .666$\pm$.025 & .655$\pm$.064 & .617$\pm$.067 & .650$\pm$.059\\
\hline
40 & .755$\pm$.039 & .676$\pm$.016 & .683$\pm$.034 & .861$\pm$.031 & .662$\pm$.054 \\
\hline
60& \textbf{.768$\pm$.029} & \textbf{.685$\pm$.011} & \textbf{.694$\pm$.040} & \textbf{.939$\pm$.020} & \textbf{.690$\pm$.040}\\
\hline
        \end{tabular}
        
\end{table}

\subsubsection{Storage Fit}
Figure~\ref{figure:buffer_size_impact} and Table~\ref{table:larger buffer better performance} provides the performance comparisons between different buffer sizes from both the perspective of numerical values and figure. We can see that larger buffer brings better performance which validates Theorem~\ref{thm:unbiased}. With this regard, our method SF$^2$EL can fit for different storages to maximize the performance by taking full advantage of the budget. We can also see that the Swiss dataset which possesses the best manifold property enjoys most the increasing of the buffer size.

\section{Conclusion}
\label{section:conclusion}
Learning with feature evolvable streams usually assumes that a label can be revealed immediately in each round. However, in reality this assumption may not hold. We introduce manifold regularization into FESL and let FESL work well in this scenario. Other feature evolvable learning can also adapt to our framework. Both theoretical and experimental results validate that our method can follow the best baselines and thus work well at any time step. Besides, we theoretically and empirically demonstrate that a larger buffer can bring better performance and thus our method can fit for different storages by taking full advantage of the budget. 


\section*{Acknowledgements}
We would like to thank all the reviewers for their helpful comments and thank Zhi-Hao Tan for valuable discussions. 


\bibliography{SF2EL}

\begin{thebibliography}{21}
\providecommand{\natexlab}[1]{#1}
\providecommand{\url}[1]{\texttt{#1}}
\providecommand{\urlprefix}{URL }
\expandafter\ifx\csname urlstyle\endcsname\relax
  \providecommand{\doi}[1]{doi:\discretionary{}{}{}#1}\else
  \providecommand{\doi}{doi:\discretionary{}{}{}\begingroup
  \urlstyle{rm}\Url}\fi

\bibitem[{Beyazit, Alagurajah, and Wu(2019)}]{DBLP:conf/aaai/BeyazitA019}
Beyazit, E.; Alagurajah, J.; and Wu, X. 2019.
\newblock Online Learning from Data Streams with Varying Feature Spaces.
\newblock In \emph{Proceedings of the 33rd {AAAI} Conference on Artificial
  Intelligence}, 3232--3239.

\bibitem[{Cesa-Bianchi and Lugosi(2006)}]{DBLP:books/daglib/0016248}
Cesa-Bianchi, N.; and Lugosi, G. 2006.
\newblock \emph{Prediction, Learning, and Games}.
\newblock Cambridge University Press.

\bibitem[{Goldberg, Li, and Zhu(2008)}]{DBLP:conf/pkdd/GoldbergLZ08}
Goldberg, A.~B.; Li, M.; and Zhu, X. 2008.
\newblock Online Manifold Regularization: {A} New Learning Setting and
  Empirical Study.
\newblock In \emph{Proceedings of the 19th European Conference on Machine
  Learning and Principles of Knowledge Discovery in Databases}, 393--407.

\bibitem[{He et~al.(2019)He, Wu, Wu, Beyazit, Chen, and
  Wu}]{DBLP:conf/ijcai/HeWWBC019}
He, Y.; Wu, B.; Wu, D.; Beyazit, E.; Chen, S.; and Wu, X. 2019.
\newblock Online Learning from Capricious Data Streams: {A} Generative
  Approach.
\newblock In \emph{Proceedings of the 28th International Joint Conference on
  Artificial Intelligence}, 2491--2497.

\bibitem[{Hou, Zhang, and Zhou(2017{\natexlab{a}})}]{DBLP:conf/nips/Hou0Z17}
Hou, B.-J.; Zhang, L.; and Zhou, Z.-H. 2017{\natexlab{a}}.
\newblock Learning with Feature Evolvable Streams.
\newblock In \emph{Advances in Neural Information Processing Systems 30},
  1417--1427.

\bibitem[{Hou, Zhang, and Zhou(2017{\natexlab{b}})}]{DBLP:conf/ijcai/Hou0Z17}
Hou, B.-J.; Zhang, L.; and Zhou, Z.-H. 2017{\natexlab{b}}.
\newblock Storage Fit Learning with Unlabeled Data.
\newblock In \emph{Proceedings of the 26th International Joint Conference on
  Artificial Intelligence}, 1844--1850.

\bibitem[{Hou, Zhang, and Zhou(2019)}]{DBLP:journals/corr/abs-1904-12171}
Hou, B.-J.; Zhang, L.; and Zhou, Z.-H. 2019.
\newblock Prediction with Unpredictable Feature Evolution.
\newblock \emph{CoRR} abs/1904.12171.

\bibitem[{Hou and Zhou(2018)}]{DBLP:journals/pami/HouZ18}
Hou, C.; and Zhou, Z.-H. 2018.
\newblock One-Pass Learning with Incremental and Decremental Features.
\newblock \emph{{IEEE} Transactions on Pattern Analysis and Machine
  Intelligence} 40(11): 2776--2792.

\bibitem[{Jiang(2008)}]{jiang2008literature}
Jiang, J. 2008.
\newblock A literature survey on domain adaptation of statistical classifiers.
\newblock \emph{URL: http://sifaka. cs. uiuc.
  edu/jiang4/domainadaptation/survey} 3: 1--12.

\bibitem[{Pan and Yang(2010)}]{DBLP:journals/tkde/PanY10}
Pan, S.~J.; and Yang, Q. 2010.
\newblock A Survey on Transfer Learning.
\newblock \emph{{IEEE} Transactions on Knowledge and Data Engineering} 22:
  1345--1359.

\bibitem[{Sch{\"{o}}lkopf and Smola(2002)}]{DBLP:books/lib/ScholkopfS02}
Sch{\"{o}}lkopf, B.; and Smola, A.~J. 2002.
\newblock \emph{Learning with Kernels: support vector machines, regularization,
  optimization, and beyond}.
\newblock Adaptive computation and machine learning series. {MIT} Press.

\bibitem[{Sun, Shi, and Wu(2015)}]{DBLP:journals/inffus/SunSW15}
Sun, S.; Shi, H.; and Wu, Y. 2015.
\newblock A survey of multi-source domain adaptation.
\newblock \emph{Information Fusion} 24: 84--92.

\bibitem[{Vincent and Bengio(2002)}]{DBLP:journals/ml/VincentB02}
Vincent, P.; and Bengio, Y. 2002.
\newblock Kernel Matching Pursuit.
\newblock \emph{Machine Learning} 48(1-3): 165--187.

\bibitem[{Vitter(1985)}]{DBLP:journals/toms/Vitter85}
Vitter, J.~S. 1985.
\newblock Random Sampling with a Reservoir.
\newblock \emph{ACM Transactions on Mathematical Software} 11(1): 37--57.

\bibitem[{Ye et~al.(2018)Ye, Zhan, Jiang, and Zhou}]{DBLP:conf/icml/YeZ0Z18}
Ye, H.-J.; Zhan, D.-C.; Jiang, Y.; and Zhou, Z.-H. 2018.
\newblock Rectify Heterogeneous Models with Semantic Mapping.
\newblock In \emph{Proceedings of the 35th International Conference on Machine
  Learning}, 1904--1913.

\bibitem[{Zhang et~al.(2016)Zhang, Zhang, Long, Ding, Zhang, and
  Wu}]{DBLP:journals/tkde/ZhangZL0ZW16}
Zhang, Q.; Zhang, P.; Long, G.; Ding, W.; Zhang, C.; and Wu, X. 2016.
\newblock Online Learning from Trapezoidal Data Streams.
\newblock \emph{{IEEE} Transactions on Knowledge and Data Engineering} 28:
  2709--2723.

\bibitem[{Zhang et~al.(2020)Zhang, Zhao, Jiang, and
  Zhou}]{DBLP:conf/icml/Zhang0JZ20}
Zhang, Z.-Y.; Zhao, P.; Jiang, Y.; and Zhou, Z.-H. 2020.
\newblock Learning with Feature and Distribution Evolvable Streams.
\newblock In \emph{Proceedings of the 37th International Conference on Machine
  Learning}, 11317--11327.

\bibitem[{Zhao et~al.(2014)Zhao, Hoi, Wang, and
  Li}]{DBLP:journals/ai/ZhaoHWL14}
Zhao, P.; Hoi, S.; Wang, J.; and Li, B. 2014.
\newblock Online Transfer Learning.
\newblock \emph{Artificial Intelligence} 216: 76--102.

\bibitem[{Zhou et~al.(2009)Zhou, Ng, She, and
  Jiang}]{DBLP:conf/pakdd/ZhouNSJ09}
Zhou, Z.-H.; Ng, M.~K.; She, Q.-Q.; and Jiang, Y. 2009.
\newblock Budget Semi-supervised Learning.
\newblock In \emph{Proceedings of 13th Pacific-Asia Conference on Knowledge
  Discovery and Data Mining}, 588--595.

\bibitem[{Zhu, Lafferty, and Rosenfeld(2005)}]{zhu2005semi}
Zhu, X.; Lafferty, J.; and Rosenfeld, R. 2005.
\newblock \emph{Semi-supervised learning with graphs}.
\newblock Ph.D. thesis, Carnegie Mellon University, language technologies
  institute, school of computer science.

\bibitem[{Zinkevich(2003)}]{DBLP:conf/icml/Zinkevich03}
Zinkevich, M. 2003.
\newblock Online Convex Programming and Generalized Infinitesimal Gradient
  Ascent.
\newblock In \emph{Proceedings of the 20th International Conference on Machine
  Learning}, 928--936.

\end{thebibliography}

\newpage
\section{Appendix: Supplementary Materials}




In the supplementary materials, we will prove the two theorems presented in the section ``Our Approach'', and provide the details of the datasets that used in our paper and additional experimental results on two large datasets.

\subsection{Analysis}

In this section, we will give the detailed proofs of the two theorems in the section “Our Approach”.

\subsubsection{Proof of Theorem~\ref{theorem:SF2EL}}
\begin{proof}
In order to prove Theorem~\ref{theorem:SF2EL}, we first introduce potential function $\Phi_t=\frac{1}{\eta}\ln\left(\sum_{i=1}^2\exp(-\eta \mathcal{J}_{i,t})\right)$. Then we have 
\begin{align*}
    \Phi_t-\Phi_{t-1}&=\frac{1}{\eta}\ln\left(\frac{\sum_{i=1}^2\exp(-\eta \mathcal{J}_{i,t})}{\sum_{i=1}^2\exp(-\eta \mathcal{J}_{i,t-1})}\right)\\
    &=\frac{1}{\eta}\left(\sum_{i=1}^2\alpha_{i,t}\exp(-\eta J_{i,t})\right)\\
    &\quad\quad(e^{-y}\leq 1-y+y^2\,\text{for all}\,y\geq 0)\\
    &\leq\frac{1}{\eta}\left(\sum_{i=1}^2\alpha_{i,t}(1-\eta J_{i,t}+\eta^2 J_{i,t}^2)\right)\\
    &=\frac{1}{\eta}\ln\left(1-\eta\sum_{i=1}^2\alpha_{i,t}J_{i,t}+\eta^2\sum_{i=1}^2\alpha_{i,t}J_{i,t}^2\right)\\
    &\quad\quad(\ln(1+y)\leq y)\\
    &\leq-\sum_{i=1}^2\alpha_{i,t}J_{i,t}+\eta\sum_{i=1}^2\alpha_{i,t}J_{i,t}^2
\end{align*}

Summing over $t=T_1+1,\ldots,T_1+T_2$, letting $\mathcal{J}_{i,T_1}=0$, telescoping and rearranging give
\begin{align*}
    \sum_{t=T_1+1}^{T_1+T_2}\alpha_{i,t}J_{i,t}&\leq\Phi_{T_1}-\Phi_{T_1+T_2}+\eta\sum_{t=T_1+1}^{T_1+T_2}\alpha_{i,t}J_{i,t}^2\\
    &\leq\frac{\ln2}{\eta}-\frac{1}{\eta}\ln(\exp(-\eta\mathcal{J}_{i^\star,T_1+T_2}))\\
    &\quad\quad+\eta\sum_{t=T_1+1}^{T_1+T_2}\sum_{i=1}^2\alpha_{i,t}J_{i,t}^2\\
    &\leq\frac{\ln2}{\eta}+\mathcal{J}_{i^\star,T_1+T_2}+\eta\sum_{t=T_1+1}^{T_1+T_2}\sum_{i=1}^2\alpha_{i,t}J_{i,t},
\end{align*}
where \[\sum_{t=T_1+1}^{T_1+T_2}\alpha_{i,t}J_{i,t}=\mathcal{J}^{S_{12}}\]
and \[\mathcal{J}_{i^\star,T_1+T_2}=\min(\mathcal{J}^{S_1},\mathcal{J}^{S_2}).\]   Considering the boundness of $J_t$, thus we have 
\begin{equation}
    \mathcal{J}^{S_{12}}\leq \min(\mathcal{J}^{S_1},\mathcal{J}^{S_2})+\frac{\ln2}{\eta}+T_2\eta.
\end{equation}
When $\eta$ is optimally set to $\sqrt{\ln2/T_2}$, (\ref{equation:SF2EL}) can be immediately derived.
\end{proof}

\subsubsection{Proof of Theorem~\ref{thm:unbiased}}
\begin{proof}
The proof of this theorem can be derived simply by using induction. \\
Basis:

$i=k: \,\textbf{Pr}(\x_t\in\mathcal{B}_t)=\frac{b}{t}$, where $\mathcal{B}_t$ is the $\mathcal{B}$ in $t$-th step.
\\
Induction $t\rightarrow t+1$:
\begin{itemize}
    \item Assume our first $t$ elements have been chosen with probability $\frac{b}{t}$.
    \item The algorithm chooses the $i+1$-th element with probability $\frac{b}{t+1}$.
    \item If we choose this element, each element in $\mathcal{B}_t$ has probability $\frac{1}{b}$ of being replaced.
    \item The probability that an element in $\mathcal{B}_t$ is replaced with the $t+1$-th element is therefore $\frac{1}{b}\frac{b}{t+1}=\frac{1}{t+1}$. 
    \item Thus the probability of an element not being replaced is $1-\frac{1}{t+1}=\frac{t}{t+1}$. 
    \item So $\mathcal{B}_{t+1}$ contains any given element either because it was chosen into $\mathcal{B}_{t}$ and not replaced: $\textbf{Pr}(\x_t\in\mathcal{B}_{t+1})=\frac{b}{t}\frac{t}{t+1}=\frac{b}{t+1}$. 
    \item Or because it was chosen in the latest round with probability $\frac{b}{t+1}$. 
\end{itemize}
We already prove that each element in $\mathcal{B}$ at time step $t$ is sampled with equal quality $p_{s\in\mathcal{B}}=\frac{b}{t}$. Then we have
\begin{align*}
\mathbb{E}[R_t]&=\sum_{s=1}^{t-1}\mathbb{E}[r_s]=\sum_{s=1}^{t-1}\sum_{k\in\mathcal{B}_t} p_k r_k\\
&=(t-1)\sum_{k\in\mathcal{B}_t} p_k r_k=\frac{t-1}{b}b\sum_{k\in\mathcal{B}_t} p_k r_k\\
&=\frac{t-1}{b}\sum_{s\in\mathcal{B}_t}\sum_{k\in\mathcal{B}_t} p_k r_k=\frac{t-1}{b}\sum_{s\in\mathcal{B}_t}\mathbb{E}[r_s]\\
&=\mathbb{E}\left[\frac{t-1}{b}\sum_{s\in\mathcal{B}_t}r_s\right]=\mathbb{E}[\hat{R}_t]
\end{align*}
where $r_s=f_{i,t}(\x_s^{S_i})-f_{i,t}(\x_t^{S_i}))^2w_{st},\,i=1,2$.

\end{proof}

\subsection{Datasets}
In the following, we present the details of all the datasets we used in our paper. The information of dimensions is presented in Table~\ref{table:dimension} where $n$ is the number of instances, $d_1$ is the number of features of the feature space $S_1$, and $d_2$ is the number of features of the feature space $S_2$. It is worth noting that RFID is a real dataset.

\begin{table}[!h]
    \renewcommand\arraystretch{1.35}
    \caption{\small The dimensions of the datasets.}
    \label{table:dimension}
    \centering
    \small
    \setlength\tabcolsep{15pt}
    \begin{tabular}{l|c|c|c}
    \hline
        \makecell[c]{Dataset} & \makecell[c]{$n$}  & \makecell[c]{$d_1$} & \makecell[c]{$d_2$}\\  
        \hline
        Credit-a &	653 &	15 &	10 \\
        Diabetes &	768	&    8 &	5 \\
        Svmguide3 & 1,284 &	22 &	15\\
        Swiss &	2000 &	2	& 3 \\
        RFID &	940	 & 78	& 72 \\
        HTRU\_2 & 17898 & 8 & 5 \\
        magic04 & 19020 & 10 & 7 \\
        \hline
    \end{tabular}
\end{table}

The descriptions and the sources of the datasets in our experiment are as follows:
\begin{itemize}
    \item Credit\_a: this dataset concerns credit card applications. All attribute names and values have been changed to meaningless symbols to protect confidentiality of the data. Credit\_a can be found in \url{https://archive.ics.uci.edu/ml/datasets/Credit+Approval}.
    \item Diabetes: this dataset consists of diabetes patients data and is from \url{https://archive.ics.uci.edu/ml/datasets/diabetes}.
    \item Svmguide3: this dataset is from \citep{hsu2003practical} and can be accessed via \url{https://www.csie.ntu.edu.tw/~cjlin/libsvmtools/datasets/binary.html}.
    \item Swiss: this dataset is a synthetic dataset containing 2000 samples and is generated by two twisted spiral datasets. As Swiss has only two dimensions, it is convenient for us to observe its manifold characteristic.
    \item RFID: this dataset is a real dataset from \citep{DBLP:conf/nips/Hou0Z17}. The authors used the RFID technique to collect the real data which contains 450 samples from $S_1$ and $S_2$ respectively. In this dataset, the RFID technique is used to predict the location's coordinate of the moving goods attached by RFID tags. RFID dataset is from \url{http://www.lamda.nju.edu.cn/data_RFID.ashx}.
    \item HTRU\_2: this dataset describes a sample of pulsar candidates collected during the High Time Resolution Universe Survey (South). It can be obtained from \url{https://archive.ics.uci.edu/ml/datasets/HTRU2}.
    \item magic04: the data are MC generated to simulate registration of high energy gamma particles in a ground-based atmospheric Cherenkov gamma telescope using the imaging technique. You can get this dataset from \url{https://archive.ics.uci.edu/ml/datasets/magic+gamma+telescope}.
\end{itemize}

\subsection{Additional Experiments}
In this section, we provide the additional experimental results on the two large datasets HTRU\_2 and magic04.

Note that we have three claims mentioned in Introduction. The first is that our method can always follow the best baseline at any time and thus achieve the fundamental goal of feature evolvable learning: always keeps the performance at a good level. The second is that manifold regularization brings better performance when there are only a few labels. The last one is that larger buffer will bring better performance and thus our method can fit for different storages by taking full advantage of the budget. 

\begin{figure}[!h]
\centering
\small
\begin{minipage}{0.45\linewidth}\centering
    \includegraphics[width=1\textwidth]{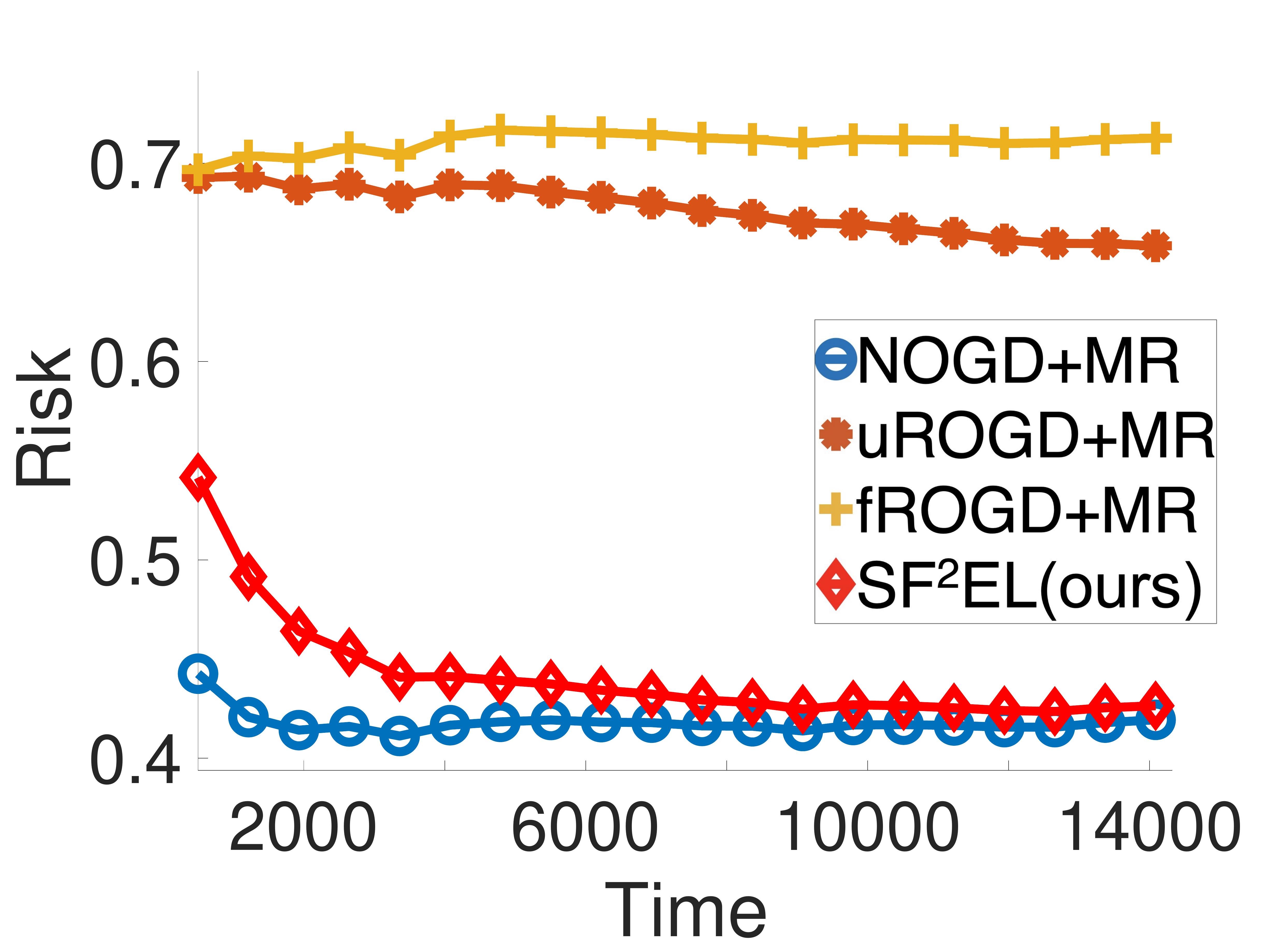}\\
    \mbox{\scriptsize\quad\quad(a) \emph{HTRU\_2}}
\end{minipage}
\begin{minipage}{0.45\linewidth}\centering
    \includegraphics[width=1\textwidth]{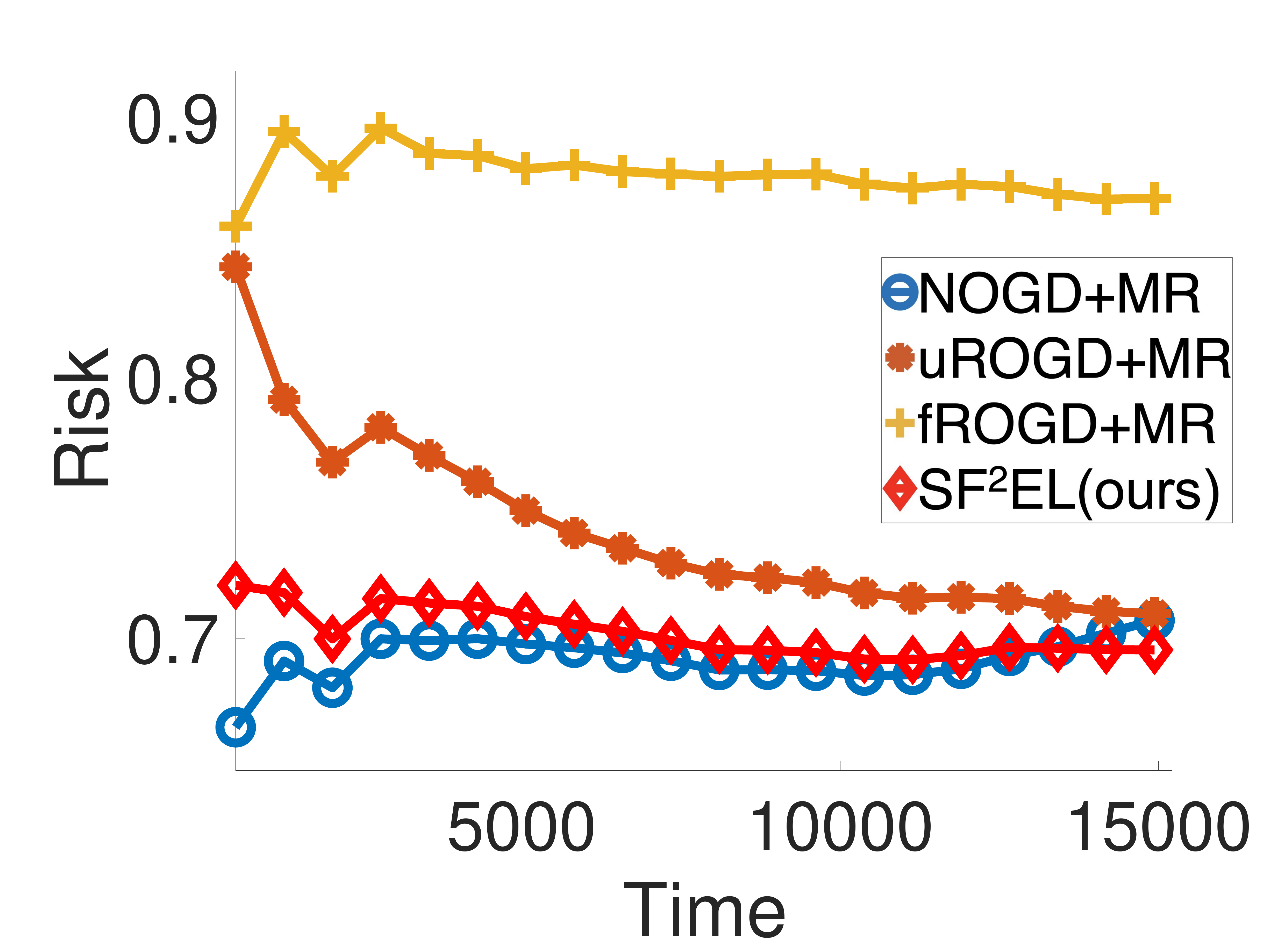}\\
    \mbox{\scriptsize\quad\quad(b) \emph{magic04}}
\end{minipage}
\vspace{-0.4em}
\caption{\small{The trend of risk with NOGD+MR, uROGD+MR, fROGD+MR and SF$^2$EL on large-scale datasets HTRU\_2 and magic04. The smaller the cumulative risk is, the better. All the average cumulative risk at any time of our method is comparable to the best baselines.}}
\label{figure:performance comparisons large data}
\vspace{-0.2cm}
\end{figure}

The numerical accuracy comparisons of the two large datasets have been exhibited in Table~\ref{table:performance comparisons} in the main body. Figure~\ref{figure:performance comparisons large data} shows the trend of risk of our method and the baselines boosted by manifold regularization. We only compare SF$^2$EL with baselines boosted by MR because those without MR cannot calculate a risk when there is no label. The smaller the cumulative risk is, the better. Similar with the results illustrated in Figure~\ref{figure:performance comparisons} in the main body, on these two large datasets, fROGD+MR's risk sometimes increases because it does not update itself.
Note that our goal is let our model be comparable to the best baseline yet is not necessary to be better than them. We can see that our method's risk on the two large datasets is always comparable with the best baseline which validates Theorem~\ref{theorem:SF2EL}.

\begin{table}[!h]
\renewcommand\arraystretch{1.5}
    \caption{\small Accuracy (mean$\pm$std) comparisons with different buffer sizes on large-scale datasets HTRU\_2 and magic04. The best ones among all the buffers are bold. We can find that larger buffer brings better performance.}
    \label{table:larger buffer better performance large data}
    \centering
    \small
    \setlength\tabcolsep{15pt}
        \begin{tabular}{c|c|c}
    \hline
    \makecell[c]{Buffer} & \makecell[c]{HTRU\_2} & \makecell[c]{magic04}\\  
    \hline
    10 & .890$\pm$.029 & .559$\pm$.053 \\
    \hline
    20 & .906$\pm$.003 & .589$\pm$.036 \\
    \hline
    40 & .907$\pm$.002 & .601$\pm$.061 \\
    \hline
    60& \textbf{.912$\pm$.008} & \textbf{.641$\pm$.026}\\
    \hline
    \end{tabular}
\end{table}

\begin{figure}[!h]
    \begin{minipage}{0.45\linewidth}\centering
     \includegraphics[width=\textwidth]{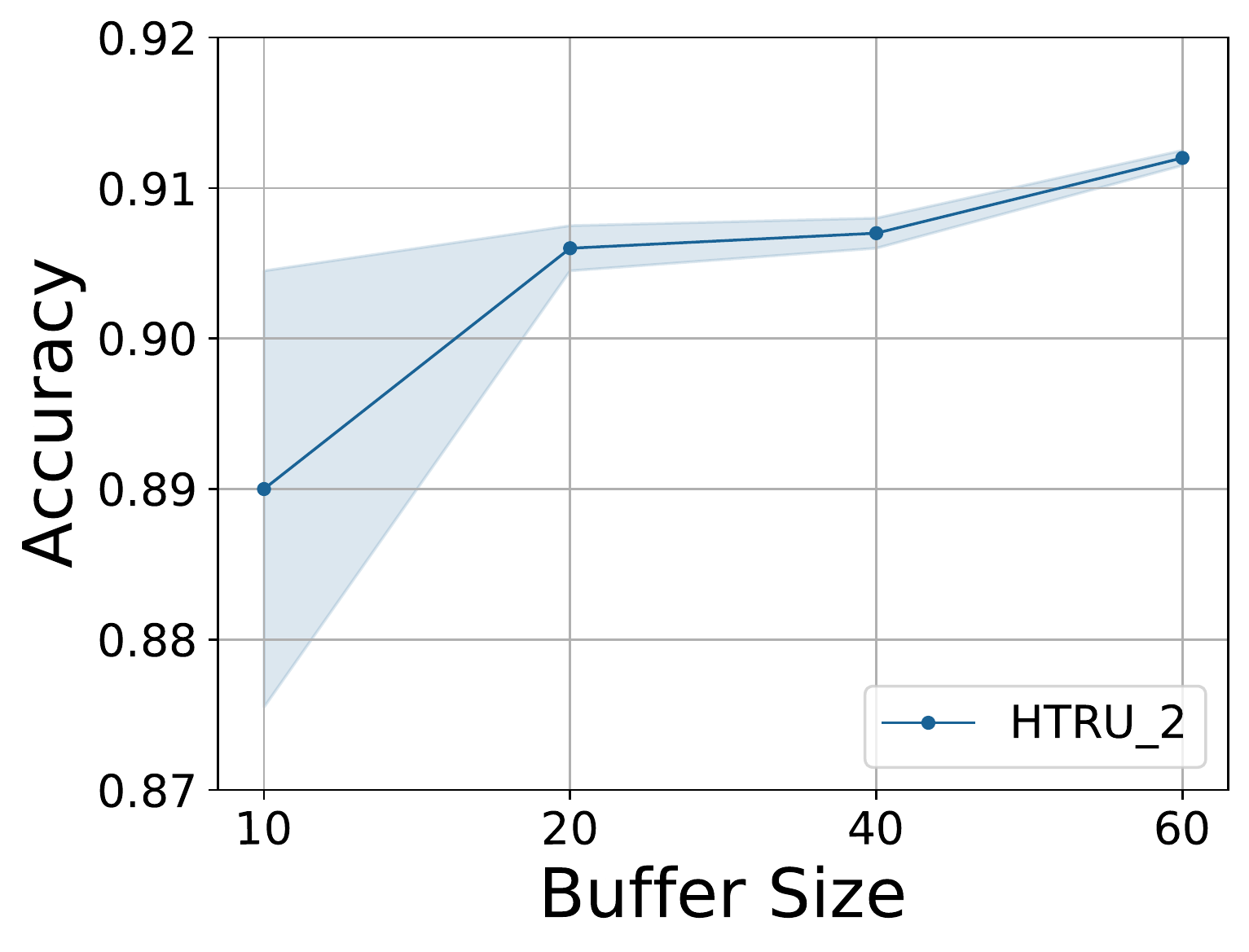}
     \mbox{\scriptsize\quad\quad(a) \emph{HTRU\_2}}
     \end{minipage}
    \begin{minipage}{0.45\linewidth}\centering
     \includegraphics[width=\textwidth]{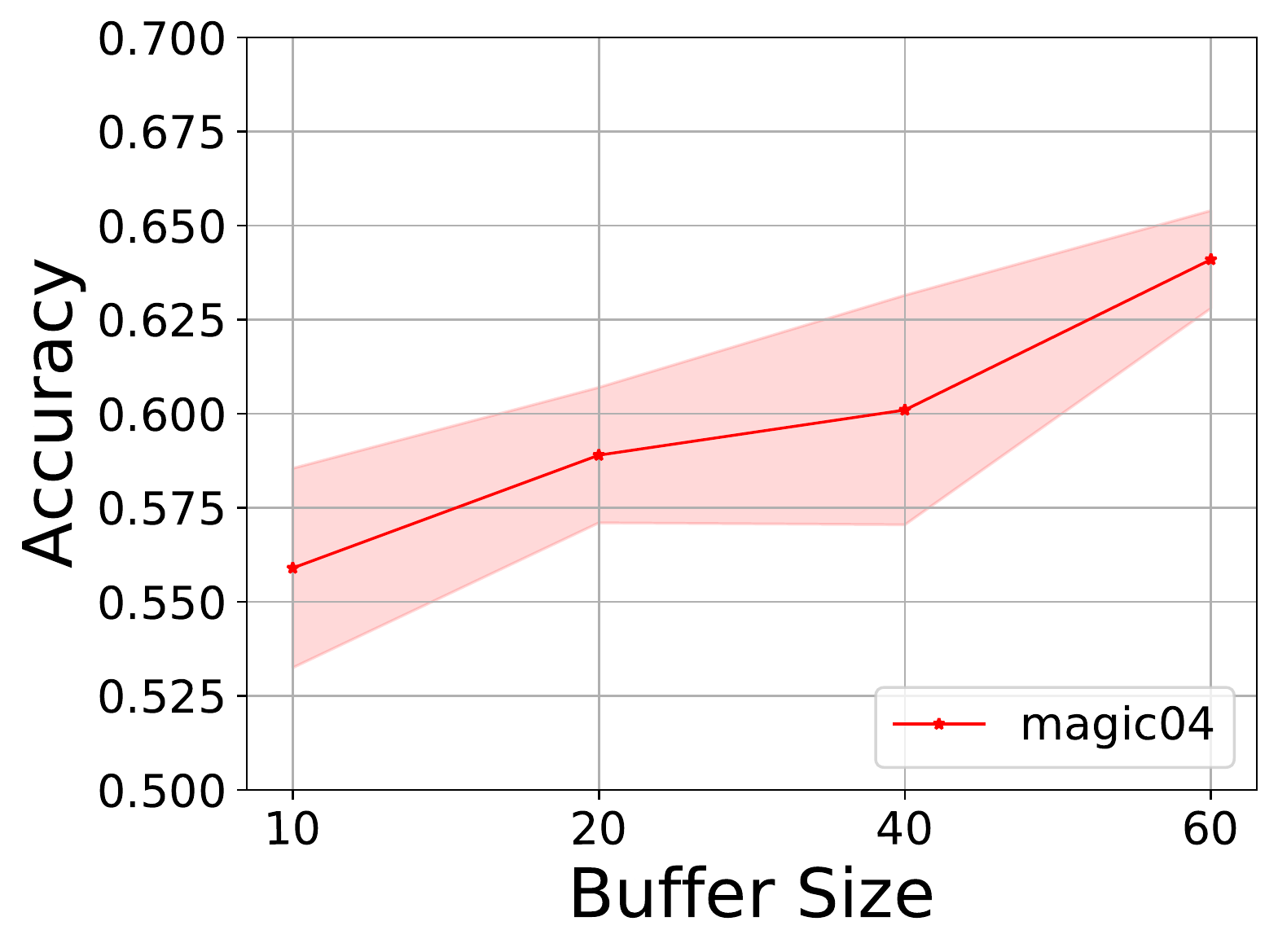}
     \mbox{\scriptsize\quad\quad(b) \emph{magic04}}
     \end{minipage}
  \vspace{0cm}
  \caption{\small The impact of buffer size on accuracy on large datasets HTRU\_2 and magic04.}
 \label{figure:buffer_size_impact large dataset}
    \vspace{-3.5mm}
\end{figure}

Table~\ref{table:larger buffer better performance large data} and Figure~\ref{figure:buffer_size_impact large dataset} provide the performance comparisons between different buffer sizes from both the perspective of numerical values and figure. We can see that similar with the results shown in Table~\ref{table:larger buffer better performance} and Figure~\ref{figure:buffer_size_impact} in the main body, larger buffer brings better performance which validates Theorem~\ref{thm:unbiased}. With this regard, our method SF$^2$EL can fit for different storages to maximize the performance by taking full advantage of the budget.



\end{document}